\PassOptionsToPackage{sort&compress}{natbib}
\documentclass[]{bytedance_seed}

\usepackage[toc,page,header]{appendix}

\usepackage{minitoc}
\usepackage{amsmath}
\usepackage{amssymb}
\usepackage{float}
\floatstyle{ruled}
\newfloat{algorithm}{t!}{loa}
\floatname{algorithm}{Algorithm}
\usepackage{algpseudocode}
\usepackage{colortbl}
\usepackage{subcaption}
\usepackage{tikz}
\usetikzlibrary{arrows.meta,calc,positioning}

\providecommand{\shortopd}{ShortOPD}
\providecommand{\best}[1]{{\bfseries #1}}
\providecommand{\cmark}{\textcolor{teal}{\checkmark}}
\providecommand{\xmark}{\textcolor{red!70!black}{\ensuremath{\times}}}
\newcommand\blfootnote[1]{%
  \begingroup
  \renewcommand\thefootnote{}\footnote{#1}%
  \addtocounter{footnote}{-1}%
  \endgroup
}

\fancypagestyle{firststyle}{
  \fancyhead[L]{\vskip 4mm
    \includegraphics[width=35mm,height=6mm]{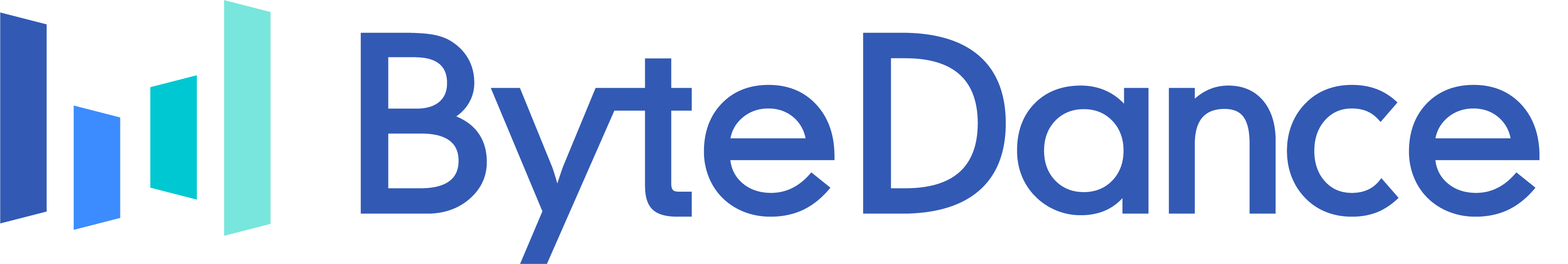}}
  \fancyhead[R]{\vskip 4mm
    {\sffamily\itshape\color{gray!60!black}Work in Progress}}
}

\title{ShortOPD: Recovering Pruned LLMs with Short-to-Long On-Policy Distillation}

\author[2,3]{Qingyu Zhang}
\author[2,3]{Qianhao Yuan}
\author[2,\dagger]{Hongyu Lin}
\author[2]{\\Yaojie Lu}
\author[2,3]{Xianpei Han}
\author[2,3]{Le Sun}
\author[1]{Ming Xu}
\author[1]{Jiarui Li}

\affiliation[1]{ByteDance}
\affiliation[2]{Chinese Information Processing Laboratory, Institute of Software, Chinese Academy of Sciences}
\affiliation[3]{University of Chinese Academy of Sciences}

\contribution[\dagger]{Corresponding author}

\abstract{
Structured pruning is a hardware-friendly way to compress LLMs, but it is
mostly validated on multiple-choice recognition tasks, while the same
compressed checkpoints can collapse on the free-form generation that
deployment actually requires. Two observations trace this gap. First,
greedy \textsc{pass}@$1$ nearly vanishes after compression, yet
\textsc{pass}@$k$ recovers substantially under repeated sampling: useful
generations are demoted, not erased. Second, in the recoverable regime
we study, failure manifests mainly as suffix repetition. Recovery should therefore train on the compressed
model's own on-policy states with dense token-level supervision, which
On-Policy Distillation (OPD) provides by reusing the pre-compression
model as a frozen teacher. However, long on-policy rollouts spend early
recovery budget on low-information repetitive suffixes, delaying loss
descent. To mitigate this waste, we propose \textbf{\shortopd}, a short-to-long OPD schedule that
detects teacher-confirmed repetitive suffixes, treats the surviving
prefix as each rollout's effective length, and allocates future rollout
budgets to the effective lengths the policy can currently use. Across
math, code, and open-ended generation, \shortopd\ raises the compressed
model's score to about $9\times$ its unrecovered value and
$1.6$--$4.4\times$ standard recovery recipes (SFT w/o KD, KD, and SeqKD),
and it matches a fixed $8192$-token rollout horizon within two points
using a quarter of the training time ($8.5$ vs.\ $35.9$ hours) and $71\%$
fewer rollout tokens.
We hope this recipe helps move structured pruning beyond marginal gains
on perplexity and multiple-choice benchmarks, a step closer to
deployment-ready generation quality.
}

\begin{document}
\maketitle

% page-1 teaser: headline recovery bars, pinned below the abstract
% (minipage + \captionof instead of a float, to control spacing exactly)
\vspace{-5mm}
\noindent\begin{minipage}{\linewidth}
\centering
\includegraphics[width=0.96\linewidth]{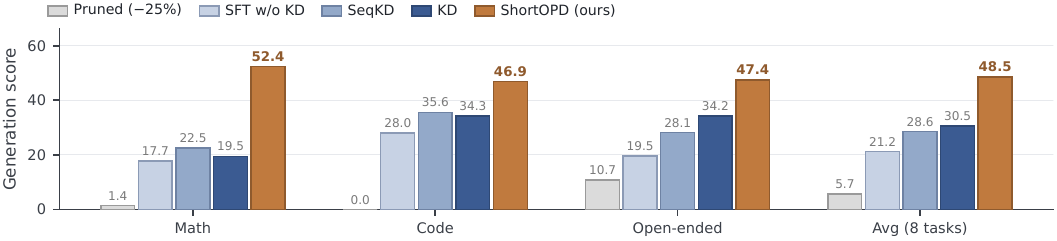}
\captionof{figure}{Headline recovery on $25\%$-pruned Qwen3-4B-Instruct:
domain-mean generation scores from Table~\ref{tab:main-results}
(open-ended judge scores $\times 10$). \shortopd\ restores about
two-thirds of the unpruned teacher's average of $75.2$.}
\label{fig:intro-headline}
\end{minipage}

%不需要目录就注释掉 注意目录不要和第一页放在一块 要有\newpage
%\newpage
%\tableofcontents
%\newpage

\section{Introduction}
\label{sec:intro}

Serving LLMs at full size is expensive~\citep{grattafiori2024llama3,qwen3},
and structured pruning is one of the most deployment-friendly ways to cut
this cost: it removes coupled parameter blocks while preserving ordinary
dense transformer execution~\citep{ma2023llmpruner,ashkboos2024slicegpt,men2025shortgpt,xia2024sheared,muralidharan2024minitron},
with no need for the dedicated kernel or hardware support that
unstructured sparsity and low-bit quantization lean
on~\citep{frantar2023sparsegpt,sun2024wanda,lin2024awq}.
The obstacle is evaluation: most pruning
work~\citep{ma2023llmpruner,ashkboos2024slicegpt,men2025shortgpt,an2024flap,kim2024shortened,gromov2025unreasonable,muralidharan2024minitron}
reports strong retention on multiple-choice recognition benchmarks such
as MMLU and HellaSwag~\citep{hendrycks2021mmlu,zellers2019hellaswag}, yet
the same
compressed checkpoints collapse on the free-form generation that
deployment actually requires~\citep{he2026demystifying,wang2025fewer,shrestha2026limits}. Removing just $4$ of $36$ layers from
Qwen3-4B-Instruct~\citep{qwen3} with Block-Influence depth
pruning~\citep{men2025shortgpt} already devastates greedy generation
across math, code, and open-ended tasks
(Table~\ref{tab:intro-passk}, \textsc{pass}@$1$ versus teacher greedy). This
recognition-generation gap is what blocks practical
use of structured pruning.\blfootnote{The experiments in this paper can
be reproduced with public implementations of the two building blocks:
structured pruning follows \url{https://github.com/icip-cas/ShortX},
and on-policy distillation follows
\url{https://github.com/VisionOPD/Vision-OPD}.}

\begin{table}[t!]
\centering
\small
\renewcommand{\arraystretch}{1.12}
\setlength{\tabcolsep}{4.5pt}
\caption{Pruning collapses generation, but correct outputs stay reachable
by sampling: best-of-$k$ scores of Qwen3-4B-Instruct with $4$ of $36$
layers removed, against the number of sampled attempts $k$. Math/code
report \textsc{pass}@$k$ (\%); open-ended tasks report LLM-judge
best-of-$k$ scores on a $1$--$10$ scale. Teacher is the unpruned model's
greedy score. \textbf{Bold}: GSM8K crosses the teacher's greedy score under
sampling---correct trajectories are demoted, not erased.}
\label{tab:intro-passk}
\begin{tabular}{llcccccccc}
\toprule
 & & \multicolumn{7}{c}{Pruned ($-4$ layers), best-of-$k$} & \\
\cmidrule(lr){3-9}
Domain & Benchmark & $k{=}1$ & $2$ & $4$ & $8$ & $16$ & $32$ & $64$ & Teacher \\
\midrule
\multirow{2}{*}{Math}
 & GSM8K     & 49.0 & 62.7 & 72.9 & 80.1 & 85.2 & 88.6 & 91.2 & 88.1 \\
 & MATH500   & 17.7 & 25.6 & 33.4 & 40.5 & 46.9 & 53.0 & 58.6 & 89.4 \\
\midrule
\multirow{2}{*}{Code}
 & HumanEval &  2.7 &  4.1 &  6.0 &  8.4 & 11.6 & 15.4 & 20.1 & 75.0 \\
 & MBPP      &  6.6 & 10.6 & 15.2 & 20.2 & 25.3 & 30.4 & 35.0 & 80.5 \\
\midrule
\multirow{4}{*}{Open-ended}
 & Alpaca    & 2.91 & 3.74 & 4.34 & 4.76 & 5.09 & 5.39 & 5.75 & 6.64 \\
 & QA        & 0.33 & 0.56 & 0.84 & 1.14 & 1.43 & 1.76 & 2.13 & 4.38 \\
 & Sum       & 3.36 & 4.66 & 5.85 & 6.91 & 7.80 & 8.44 & 8.88 & 8.86 \\
 & MT-Bench  & 2.19 & 2.92 & 3.63 & 4.27 & 4.83 & 5.35 & 5.88 & 6.95 \\
\bottomrule
\end{tabular}
\end{table}

Two observations identify the recovery signal. First, the missing
ability is demoted rather than erased. \citet{wen2026illusion} show this
for short QA answers; we find the same holds for entire generation
trajectories: in Table~\ref{tab:intro-passk},
\textsc{pass}@$k$~\citep{chen2021codex} climbs steadily with more
sampled attempts on every benchmark, and on GSM8K~\citep{cobbe2021gsm8k}
it even reaches $91\%$ at $k{=}64$, above the unpruned teacher's greedy
$88\%$. Correct trajectories thus
remain in the compressed model's own sampling distribution, within reach
of on-policy search.
Second, the repair must be token-level: ShortGPT-gen restores much of the
lost quality by routing only \textit{generated} tokens through the full
model~\citep{men2025shortgpt}, so generation failure is a chain of
local next-token mis-rankings that compound along the student's own
decoding path~\citep{ranzato2016sequence,agarwal2024gkd,he2026demystifying}. Sparse
rewards~\citep{shao2024deepseekmath,deepseekr1} and fixed off-policy
labels~\citep{kim2016sequence} leave these local errors unspecified. Recovery
therefore needs dense distributional targets on the states the compressed
model actually visits.

\begin{figure}[t!]
\centering
\begin{subfigure}[b]{0.490\linewidth}
\centering
\includegraphics[width=\linewidth]{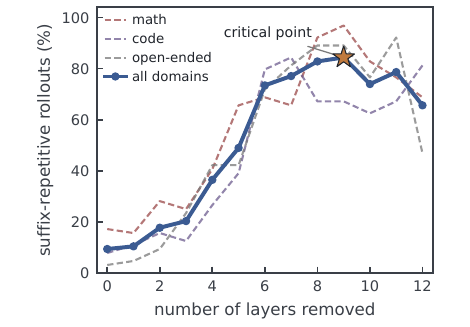}
\caption{Suffix repetition vs.\ pruned depth}
\end{subfigure}\hfill
\begin{subfigure}[b]{0.490\linewidth}
\centering
\includegraphics[width=\linewidth]{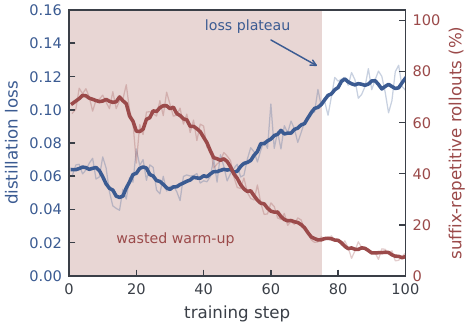}
\caption{Repetitive suffixes coincide with low loss}
\end{subfigure}
\caption{Repetition during early on-policy recovery.
\textbf{(a)} Suffix-repetition rate versus BI-pruned depth on a fixed
$192$-prompt probe; the star marks our main $9$-layer setting.
\textbf{(b)} Distillation loss (blue) and suffix-repetition rate (red)
during fixed-$H{=}2048$ OPD; shading marks the repetition-dominated
warm-up. Figure~\ref{fig:shortopd-traj}(a) overlays \shortopd.}
\label{fig:intro-prune-rep}
\end{figure}

OPD~\citep{agarwal2024gkd,yuan2026vision} is the direct instantiation of
these requirements. The compressed student samples rollouts from its own
distribution. Its frozen pre-compression self scores the same
trajectories as teacher, and the student matches the teacher's next-token
distribution at every response position. The recipe is on-policy and
dense, and it needs no labels, no verifier, and no external teacher.
The remaining difficulty is that, consistent with recent evidence of
pruning-induced reasoning loops~\citep{wang2025fewer}, early on-policy
rollouts are dominated by suffix repetition~\citep{holtzman2020degeneration,xu2022ditto,pipis2026wait,hiraoka2025repetition,duan2026circular}. Probing
BI-pruned Qwen3-4B-Instruct on a fixed $192$-prompt math/code/open-ended
set, suffix repetition reaches $84\%$ at our $25\%$-parameter testbed
(Figure~\ref{fig:intro-prune-rep}a); with stronger compression, outputs
turn incoherent instead of looped (Appendix~\ref{app:prunesweep}), so we
view $25\%$ as this model's critical point for recoverable compression
and adopt it as our main testbed. These loop tails also carry little measured
marginal distillation signal: repeated-suffix tokens have a mean teacher--student
generalized Jensen--Shannon divergence (JSD) of $0.0014$ versus $0.051$ on
ordinary tokens (about $35\times$ lower), and
a mean teacher NLL of $3\times10^{-5}$ versus $0.68$. Together with the
conditional gradient analysis in Appendix~\ref{app:gradient-derivation},
these measurements identify deep repeated suffixes as low-information under
the implemented OPD update: they consume early rollout and teacher compute
under a long, fixed rollout horizon $H$ (the per-response token budget) while
providing little additional teacher--student correction. Figure~\ref{fig:intro-prune-rep}(b) follows
the first $100$ steps of a fixed-$H{=}2048$ OPD run: during the
early phase, $55$--$75\%$ of rollouts end in suffix loops while the
distillation loss remains at its lowest level, reaching its post-warm-up
range only around step $80$ as repetition subsides.

We propose \shortopd\ (Short-to-Long On-Policy Distillation) to make OPD
spend its early budget where the supervision is useful. A lightweight
controller tracks suffix repetition, truncation, and mean effective length.
High repetition opens the shrink gate, whose target is an exponential moving
average (EMA) of effective length; low repetition and high truncation open the
growth gate. A second EMA moves the actual budget toward either target; both
the first smooths noisy batch estimates, while the second prevents abrupt
horizon changes.
Training therefore contains repetitive early rollouts
and restores long generation as recovery proceeds, at no extra forward
cost. In practice,
\shortopd\ narrows the high-repetition region, reduces tokens spent in
repeated suffixes, and improves recovery over fixed short and fixed long
horizons (Figure~\ref{fig:intro-headline}).

\noindent Our contributions are threefold:
\begin{itemize}
\item \textbf{We propose \shortopd}, a recipe for recovering structurally
compressed LLMs built on our diagnosis of what recovery requires:
supervision on the student's own on-policy states, dense token-level
teacher distributions, and a rollout budget matched to what the damaged
policy can currently generate. \shortopd\ delivers the third ingredient
with repetition-gated, truncation-aware horizon control.
\item \textbf{Extensive experiments demonstrate the effectiveness and
efficiency of this recovery recipe.} On Qwen3-4B-Instruct with a
mixed math/code/instruction research corpus,
\shortopd\ restores about two-thirds of the unpruned teacher's generation
score over eight task families, far ahead of SFT w/o KD, KD, SeqKD,
and sparse-reward RLVR under matched budgets, and it matches fixed
rollout horizons of up to $8192$ tokens within two points while
generating up to $71\%$ fewer rollout tokens; controlled comparisons
confirm that the gain requires \textit{both} on-policy states and dense
teacher distributions.
\item \textbf{The recovery corpus matters.} On-policy distillation only
repairs the states the student's own search visits: leave-one-domain-out
ablations show that removing math or code prompts sharply damages recovery
on the corresponding capabilities. Corpus coverage is therefore a
first-class design choice for post-compression recovery, not a detail.
\end{itemize}

\section{Related Work}
\label{sec:related}

\paragraph{Structured pruning and the recognition-generation gap.}
Depth pruning is among the most serving-friendly LLM compression
families: deleting whole transformer blocks preserves a standard dense
architecture, with layers selected by Block Influence
(ShortGPT~\citep{men2025shortgpt}), merged (LaCo~\citep{yang2024laco}),
or removed under other criteria~\citep{kim2024shortened,
gromov2025unreasonable,yuan2025shortv,hu2026logits}. Width pruning instead removes coupled
structures inside layers~\citep{ma2023llmpruner,ashkboos2024slicegpt,
an2024flap,dery2024everybody}; Sheared LLaMA~\citep{xia2024sheared}
learns pruning masks with continued pre-training, and
Minitron~\citep{muralidharan2024minitron,sreenivas2024minitron} combines
depth and width pruning with distillation-based retraining. One-shot
weight sparsification (SparseGPT~\citep{frantar2023sparsegpt},
Wanda~\citep{sun2024wanda}) is complementary but needs sparse-kernel
support. Across these lines, validation is dominated by
recognition-style benchmarks such as MMLU and
HellaSwag~\citep{hendrycks2021mmlu,zellers2019hellaswag}.
\citet{wen2026illusion} show why this is deceptive: pruned models can
still score multiple-choice answers yet fail to produce them, the
missing answers demoted rather than erased~\citep{he2026demystifying,wang2025fewer,shrestha2026limits}. ShortGPT-gen makes the
generative side concrete by routing only \textit{generated} tokens
through the full depth, locating the damage along the decoding
path~\citep{men2025shortgpt}. We extend the demotion view from short
answers to entire reasoning trajectories and focus on the step these
works leave open: after the layers are gone, which training signal best
restores generation?

\paragraph{Distillation and on-policy recovery signals.}
Knowledge distillation~\citep{hinton2015distilling} trains a student to
match a teacher; sequence-level KD~\citep{kim2016sequence} trains on
fixed teacher-generated responses and is therefore off-policy. A line of
work closes the resulting train-inference mismatch by distilling on
student states: imitation-learning KD~\citep{lin2020imitkd},
MiniLLM~\citep{gu2024minillm} with a reverse KL divergence objective,
divergence generalizations~\citep{wen2023fdistill,ko2024distillm}, and
generalized on-policy KD~\citep{agarwal2024gkd}. Recovery retraining
after pruning~\citep{thangarasa2025selfdata}, however, still mostly follows the off-policy recipe:
Sheared LLaMA continues pre-training on corpus
data~\citep{xia2024sheared}, and Minitron distills teacher logits on a
fixed data blend and validates on recognition
suites~\citep{muralidharan2024minitron,sreenivas2024minitron}, so
whether off-policy logit distillation repairs \textit{generation} is
left open; our KD baseline instantiates exactly this recipe for a
controlled comparison (Section~\ref{sec:main}). RL with verifiable
rewards~\citep{shao2024deepseekmath,deepseekr1,lambert2025tulu3,
yu2025dapo} is on-policy but sparse: it offers no token-level target
when a heavily compressed model rarely samples a correct answer.
Vision-OPD~\citep{yuan2026vision} introduced on-policy distillation
for multimodal models; we study the same principle after structural
compression, where the privileged teacher is the model's own
pre-compression self. All of these methods distill a \textit{healthy}
generator; none addresses a damaged student whose rollouts collapse into
repetition~\citep{wang2025fewer,pipis2026wait,hiraoka2025repetition,duan2026circular} that the self-teacher then endorses.

\section{Method}
\label{sec:approach}

\begin{figure}[t!]
\centering
\includegraphics[width=\linewidth]{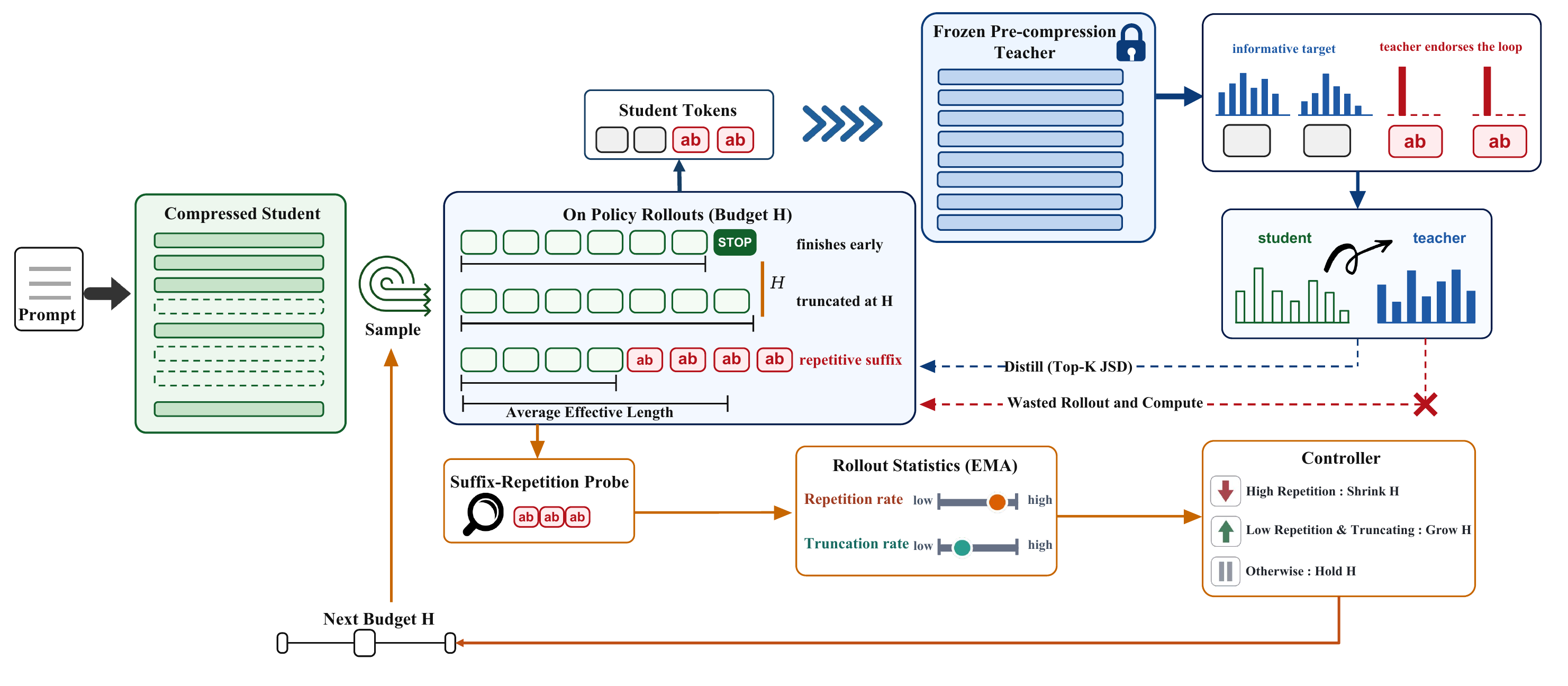}
\caption{\shortopd\ closes a second control loop around OPD.
\textbf{Distillation loop (blue):} under the current budget $H$, the
compressed student samples on-policy rollouts (finishing early, truncated
at $H$, or collapsing into a repetitive suffix); the frozen
pre-compression teacher re-scores the same tokens and returns a dense
top-$K$ distribution target at every position. \textbf{Budget loop
(orange):} a token-only terminal-periodic probe reports severe repetition;
its onset, locally refined by the existing OPD loss and teacher NLL, gives
effective length. The fraction of non-repetitive rollouts that fill $H$
gives the clean truncation rate. High repetition sets a shrink target from
the EMA of mean effective length; low repetition and high clean truncation
set a growth target. The next budget is an EMA-smoothed move toward that
target.}
\label{fig:shortopd-framework}
\end{figure}

We first formalize OPD as a recovery recipe for structurally compressed
LLMs. The experiments instantiate compression with Block-Influence depth
pruning because it is simple, hardware-friendly, and representative of
structured pruning, but the recovery objective only requires a compressed
student and its pre-compression self. We then present \shortopd, a
short-to-long OPD schedule driven by repetition and truncation feedback;
Figure~\ref{fig:shortopd-framework} gives the overview and
Algorithm~\ref{alg:shortopd} the procedure. Finally, we place \shortopd\
and its baselines in a single design space.

\subsection{Preliminaries: pruning and the OPD base recipe}
\label{sec:opd}

\paragraph{Representative structured pruning operator.}
Let $X_{i,t}$ denote the hidden state of token $t$ entering layer $i$.
ShortGPT's Block Influence (BI) of layer $i$ is
\begin{equation}
\mathrm{BI}_i \;=\; 1 - \mathbb{E}_{X,t}\!\left[
\frac{X_{i,t}^{\top} X_{i+1,t}}{\lVert X_{i,t}\rVert_2 \, \lVert X_{i+1,t}\rVert_2}
\right],
\label{eq:bi}
\end{equation}
so a low $\mathrm{BI}_i$ marks a layer that barely changes the hidden
state. We remove the lowest-BI layers until roughly $25\%$ of parameters
are gone, obtaining the compressed student $\pi_\theta$, and keep the
original model $\pi_T$ as a frozen teacher. The operator is only the
experimental object: \shortopd\ assumes nothing beyond $\pi_\theta$ being
a structurally compressed descendant of $\pi_T$.

\paragraph{On-policy self-distillation.}
Following the two implications of Section~\ref{sec:intro}, the recovery
objective should repair the student on the states it actually visits and
at token granularity. OPD satisfies both: for each prompt $x$, the
student samples on-policy rollouts $y\sim\pi_\theta(\cdot\mid x)$, the
frozen teacher $\pi_T$ is run on the same trajectories, and the student
matches the teacher's next-token distribution at every response position:
\begin{equation}
\mathcal{L}_{\mathrm{OPD}}(\theta)
= \mathbb{E}_{x}\,\mathbb{E}_{y\sim\pi_\theta(\cdot\mid x)}
\!\left[\frac{1}{|y|}\sum_{t=1}^{|y|}
D_{\alpha}\!\big(\pi_T(\cdot\mid x,y_{<t}),\,\pi_\theta(\cdot\mid x,y_{<t})\big)\right],
\label{eq:opd}
\end{equation}
where $D_{\alpha}$ is a generalized Jensen--Shannon divergence; we
distill the top-$100$ logits plus an aggregated tail mass, with clipped
importance weighting against the rollout policy. With a large
teacher--student capacity gap, forward KL is mode-covering and can force
a limited student to spread probability mass across teacher modes,
whereas reverse KL is mode-seeking but may reduce
diversity~\citep{gu2024minillm,agarwal2024gkd}. This trade-off is
independent of whether the training trajectories are off- or on-policy
and remains task-dependent~\citep{agarwal2024gkd}. We therefore use the
balanced choice $\alpha=0.5$, a bounded compromise that avoids extreme
log-ratio penalties when the teacher and damaged student differ sharply.
Equation~\eqref{eq:opd} is on-policy (the expectation is over student
samples), dense (every generated token receives a distributional
target), and reward-free, so it applies equally to verifiable math/code
prompts and open-ended instruction prompts. Because the teacher is the
pre-compression self, OPD pulls the student back toward its own original
behavior on the trajectories it actually produces.

\subsection{Short-to-Long On-Policy Distillation}
\label{sec:shortopd}

\begin{algorithm}[t!]
\caption{\shortopd}
\label{alg:shortopd}
\small
\begin{algorithmic}[1]
\Require prompts $\mathcal{B}$, student $\pi_\theta$, teacher $\pi_T$, bounds $H_{\min},H_{\max}$, thresholds $\rho_{\mathrm{low}},\rho_{\mathrm{high}},\tau$, margin $\lambda$, growth $\gamma_{\uparrow}$, EMA decays $\beta,\beta_H$
\State $H_1 \gets H_{\max}$; initialize EMAs
\For{training step $s=1,\ldots,S$}
  \State sample rollouts $y_i\sim\pi_\theta(\cdot\mid x_i)$ with budget $H_s$
  \State $a_i\gets\operatorname{TerminalLoop}(y_i)$;
  $r_s\gets B^{-1}\sum_i a_i$
  \State $q_s\gets B^{-1}\sum_i\mathbb{1}[|y_i|=H_s\ \land\ a_i=0]$
  \State run $\pi_T$ on the same trajectories
  \State update $\theta$ with the dense OPD loss over all generated tokens (Eq.~\ref{eq:opd})
  \State compute effective lengths $\ell_i$ and mean $L_s$; update $\bar r_s,\bar q_s,\bar L_s$
  \If{$\bar r_s>\rho_{\mathrm{high}}$}
    \State $H_s^*\gets\min(H_s,\lambda\bar L_s)$
  \ElsIf{$\bar r_s<\rho_{\mathrm{low}}$ \textbf{and} $\bar q_s>\tau$}
    \State $H_s^*\gets\gamma_{\uparrow}H_s$
  \Else
    \State $H_s^*\gets H_s$
  \EndIf
  \State $H_{s+1}\gets\mathrm{clip}_{[H_{\min},H_{\max}]}(\lfloor\beta_HH_s+(1-\beta_H)H_s^*\rceil_{16})$
\EndFor
\end{algorithmic}
\end{algorithm}

Equation~\eqref{eq:opd} implicitly assumes that the student's rollouts are
usable training states. Under strong compression this assumption becomes
horizon-dependent: a long fixed budget reaches repetitive suffixes before
the policy has recovered, while a permanently short budget truncates
legitimate long generations later. \shortopd\ therefore keeps the global
\texttt{response\_length} as a static padding and context ceiling, but sets
a per-step sampling budget $H_t \le H_{\max}$. The budget is not driven by
batch-mean loss, which is lowest precisely during the repetitive warm-up
(Figure~\ref{fig:intro-prune-rep}b). Instead, \shortopd\ observes three
statistics at every step: how many rollouts end in a severe periodic loop,
how many non-repetitive rollouts exhaust the current budget, and the mean
usable prefix before the detected loop. Repetition is the first-level gate:
effective length may shorten the budget only when repetition is high,
whereas low repetition and high clean truncation permit longer rollouts.

\paragraph{Generation-side feedback.}
For rollout $y_i$ at step $s$, the probe examines only its last $W$ valid
tokens. Let $a_i=1$ when the terminal-periodic detector of
Appendix~\ref{app:loop-detector} finds a severe loop and $a_i=0$ otherwise.
Let $b_i=1$ only when the rollout fills the current budget without such a
loop, i.e., $|y_i|=H_s$ and $a_i=0$. The structural loop onset is locally
refined using the OPD divergence and teacher NLL to obtain effective length
$\ell_i$; these loss signals refine the boundary but never determine
$a_i$. We set $\ell_i=|y_i|$ when no severe loop is found. The batch
statistics and their EMAs are
\begin{equation}
r_s=\frac{1}{B}\sum_{i=1}^{B}a_i,
\qquad
q_s=\frac{1}{B}\sum_{i=1}^{B}b_i,
\qquad
L_s=\frac{1}{B}\sum_{i=1}^{B}\ell_i,
\qquad
(\bar r_s,\bar q_s,\bar L_s)=\operatorname{EMA}_{\beta}(r_s,q_s,L_s).
\label{eq:controller-signals}
\end{equation}
The suffix restriction avoids reacting to repeated structure earlier in a
valid response, and the EMA suppresses single-batch fluctuations. These
statistics reuse tokens and teacher scores already produced by OPD and
require no extra forward pass. The effective-length probe is control-only:
the current batch still receives the unmodified dense OPD loss of
Equation~\eqref{eq:opd} over all generated tokens.

\paragraph{Repetition-gated horizon control.}
The controller first forms a gated target
\begin{equation}
H_s^*=
\begin{cases}
\min(H_s,\lambda\bar L_s),
    & \bar r_s>\rho_{\mathrm{high}},\\
\gamma_{\uparrow}H_s,
    & \bar r_s<\rho_{\mathrm{low}}\ \land\ \bar q_s>\tau,\\
H_s, & \text{otherwise},
\end{cases}
\qquad
H_{s+1}=\mathrm{clip}_{[H_{\min},H_{\max}]}\!\Big(\Big\lfloor
\beta_H H_s+(1-\beta_H)H_s^*
\Big\rceil_{16}\Big).
\label{eq:shortopd}
\end{equation}
Here $\lambda>1$ retains headroom beyond the observed usable prefix,
$\gamma_{\uparrow}>1$ proposes growth, and $\beta_H$ smooths both
directions before rounding to multiples of $16$. The hysteresis band
$\rho_{\mathrm{low}}<\rho_{\mathrm{high}}$ prevents oscillation. High
repetition has priority even when truncation is also high, and its severity
determines the shrink target through $\bar L_s$. Growth requires both clean
suffixes and evidence that the current horizon is binding. Consequently,
short clean responses cannot reduce the budget merely by lowering average
length. We initialize at $H_{\max}$ and checkpoint the current budget and
all three EMAs across restarts.
Algorithm~\ref{alg:shortopd} summarizes the loop.

\subsection{A design space for recovery signals}
\label{sec:design}
\shortopd\ is one cell of a space spanned by two axes: trajectory source
(on-policy student rollouts vs.\ off-policy fixed data) and supervision
(dense teacher distribution vs.\ hard labels or sparse reward). Table~\ref{tab:design}
summarizes where each recovery method sits. Our Qwen3 experiments
instantiate SFT w/o KD, SeqKD, KD, and \shortopd\ on the
same prompt distribution, and add a matched Math+Code RLVR comparison for
the sparse on-policy cell:

\begin{table}[t!]
\centering
\small
\setlength{\tabcolsep}{6pt}
\caption{Recovery methods organized by the two design axes. Only
\shortopd\ trains on the compressed student's own states with a dense
token-level signal while steering the rollout budget away from
low-information suffixes; it requires neither ground-truth responses nor a
reward verifier.}
\label{tab:design}
\begin{tabular}{lcccc}
\toprule
Method & On-policy states & Dense signal & Label-free & Verifier-free \\
\midrule
SFT w/o KD        & \xmark & \xmark & \xmark & \cmark \\
SeqKD             & \xmark & \xmark & \cmark & \cmark \\
KD                & \xmark & \cmark & \xmark & \cmark \\
RLVR              & \cmark & \xmark & \xmark & \xmark \\
\best{\shortopd\ (ours)} & \cmark & \cmark & \cmark & \cmark \\
\bottomrule
\end{tabular}
\end{table}
\begin{itemize}
  \item \textbf{SFT w/o KD}: standard supervised finetuning on the corpus's
  original golden responses. This tests whether simply training the compressed
  model on the same data is enough.
  \item \textbf{SeqKD}~\citep{kim2016sequence}: SFT on fixed responses
  generated by the same pre-compression teacher $\pi_T$. This shares OPD's teacher
  but removes the on-policy trajectory source.
  \item \textbf{KD}~\citep{hinton2015distilling,agarwal2024gkd}: teacher-forced
  token-level distillation (also known as word-level KD) that keeps SFT's fixed
  sequences but replaces hard labels with the teacher's next-token
  distributions. This shares OPD's dense signal but removes the on-policy
  trajectory source.
  \item \textbf{RLVR}~\citep{schulman2017ppo,shao2024deepseekmath}: on-policy
  learning with sparse correctness rewards on math and executable code; we
  instantiate it with both PPO and GRPO. This shares OPD's on-policy sampling
  but replaces dense teacher distributions with sparse outcome rewards.
  \item \textbf{\shortopd\ (ours)}: on-policy rollouts with dense
  self-distillation targets and the repetition-gated budget controller of
  Section~\ref{sec:shortopd}.
\end{itemize}

\section{Experiments}
\label{sec:exp}

% ============================================================================
% ============================================================================

The experiments evaluate \shortopd\ and its OPD foundation along four axes.
First, we compare recovered generation quality against practical
post-compression recovery baselines across two model scales, two generation
modes, depth/width/hybrid compression structures, and eight task families.
Second, we analyze the short-to-long horizon controller: its
trajectory, its comparison against fixed short and fixed long horizons, and
the repetition dynamics that drive it. Third, we isolate the learning
signal, separating dense on-policy distillation from off-policy SFT-style
recovery and sparse on-policy reward learning. Fourth, we test how recovery
depends on the prompt-domain mixture.

\subsection{Setup}
\label{sec:setup}

\paragraph{Backbones and compression.}
We use Qwen3-4B-Instruct-2507, Qwen3-4B-Thinking-2507, and Qwen3-8B in
both no-thinking and thinking modes~\citep{qwen3}. Each model has $36$
transformer layers. The original model is the teacher $\pi_T$, and the
compressed student is obtained by deleting the nine lowest-BI layers
($25\%$ of transformer depth). BI is computed on a held-out calibration
corpus. The teacher is frozen throughout recovery. We evaluate full
cross-domain recovery at both $4$B and $8$B. To test whether the result is
specific to layer deletion, we additionally use width-only and hybrid
width--depth pruning for both no-thinking backbones, following the ModelOpt
configurations of Minitron~\citep{muralidharan2024minitron,sreenivas2024minitron}.
The resulting students contain $3.11$B parameters at 4B scale and
$6.45$--$6.47$B parameters at 8B scale; each recovery pair starts from the
same raw pruned checkpoint.

\paragraph{Recovery corpus construction.}
We use a mixed research corpus spanning math, code, and open-ended
instruction domains. Recovery and evaluation examples are kept disjoint,
and standard quality filtering and deduplication are applied. All recovery
methods share the same prompt distribution but consume it
differently: OPD and \shortopd\ use only the prompt field, sampling
on-policy rollouts scored by the frozen teacher; SFT w/o KD trains on
the original assistant responses; SeqKD trains on fixed responses
generated by the same frozen teacher; and KD keeps SFT's fixed sequences
but replaces hard labels with the teacher's next-token distributions.
For the Thinking backbone, SFT w/o KD augments each original response with a
reasoning trace generated by the frozen teacher, while SeqKD and KD use
the teacher's complete reasoning-to-answer trajectories.
The comparisons therefore vary the recovery signal and trajectory
source, not the prompt mixture.

\paragraph{Recovery methods and schedules.}
Unless otherwise noted, all recovery methods start from the same compressed
student and use the same step budget on the same prompt distribution. KD is a teacher-forced
forward-KL baseline,
$\mathrm{KL}(\pi_T(\cdot\mid x_{<t})\|\pi_S(\cdot\mid x_{<t}))$, with
temperature $1.0$ and no hard-label cross-entropy term; it instantiates,
on our recovery corpus, the fixed-context logit-distillation recipe used
by prune-then-distill pipelines such as
Minitron~\citep{muralidharan2024minitron,sreenivas2024minitron}, so the
KD-versus-OPD contrast directly tests whether that recipe
extends from recognition benchmarks to generation recovery. For OPD, the student
samples on-policy responses; the pre-compression teacher is then evaluated on
the same trajectories, and the student applies the top-$100$+tail
generalized-JSD distillation loss of Section~\ref{sec:opd}. The fixed
horizons use a constant rollout budget ($2048$ or $8192$ tokens).
\shortopd\ uses the same loss and data with the repetition-gated budget
rule of Equation~\eqref{eq:shortopd}, initialized at $H{=}2048$ so the
budget must discover the usable range from the first step. Full hyperparameters are in
Appendix~\ref{app:train}. The nominal schedule is $710$ steps.
The reported 4B Thinking \shortopd\ checkpoint starts from the cross-domain
\shortopd\ checkpoint and then receives one additional math-domain OPD
stage ($117$ steps).
All experiments run on a single node with eight GPUs.

\begin{table}[t!]
\centering
\small
\renewcommand{\arraystretch}{1.12}
\setlength{\tabcolsep}{3.4pt}
\caption{Main generation recovery after $25\%$ depth pruning ($9/36$
layers). Math/code are percentages; open-ended columns are $1$--$10$ judge
scores. Avg rescales judge scores by $\times 10$, and T\% is relative to the
corresponding dense teacher. Decoding protocols differ across model/mode
blocks, so comparisons are within block; full settings are in
Appendix~\ref{app:eval}.}
\label{tab:main-results}
\resizebox{\linewidth}{!}{%
\begin{tabular}{llcccccccccc}
\toprule
 & & \multicolumn{2}{c}{Math} & \multicolumn{2}{c}{Code} & \multicolumn{4}{c}{Open-ended} & \multicolumn{2}{c}{Overall} \\
\cmidrule(lr){3-4}\cmidrule(lr){5-6}\cmidrule(lr){7-10}\cmidrule(lr){11-12}
Model & Method & GSM8K & MATH & HumanEval & MBPP & Alpaca & QA & Sum & MT-B & Avg & T\% \\
\midrule
\multirow{6}{*}{\shortstack[l]{Qwen3-4B\\Instruct}}
& Dense teacher             & 88.10 & 89.40 & 75.00 & 80.54 & 6.64 & 4.38 & 8.86 & 6.95 & 75.17 & 100.0 \\
& Pruned ($-25\%$)          & 1.14  & 1.60  & 0.00  & 0.00  & 1.14 & 1.00 & 1.06 & 1.09 & 5.71 & 7.6 \\
\cmidrule(lr){2-12}
& SFT w/o KD                & 25.93 & 9.40  & 26.83 & 29.18 & 1.91 & 1.19 & 3.12 & 1.59 & 21.19 & 28.2 \\
& SeqKD                     & 32.83 & 12.20 & 29.88 & 41.25 & 2.52 & 1.38 & 5.26 & 2.10 & 28.60 & 38.0 \\
& KD             & 29.34 & 9.60  & 29.27 & 39.30 & 2.94 & 1.48 & 6.89 & 2.36 & 30.52 & 40.6 \\
\rowcolor{gray!25}
& \shortopd\ (ours)         & \best{62.70} & \best{42.00} & \best{43.90} & \best{49.81} & \best{4.68} & \best{2.04} & \best{7.94} & \best{4.28} & \best{48.46} & \best{64.5} \\
\midrule
\multirow{6}{*}{\shortstack[l]{Qwen3-4B\\Thinking}}
& Dense teacher             & 92.87 & 88.80 & 76.83 & 87.16 & 5.71 & 3.40 & 8.68 & 6.65 & 73.75 & 100.0 \\
& Pruned ($-25\%$)          & 0.45 & 1.20 & 0.00 & 0.00 & 1.00 & 1.00 & 1.00 & 1.01 & 5.21 & 7.1 \\
\cmidrule(lr){2-12}
& SFT w/o KD                & 18.73 & 15.20 & 18.90 & 28.02 & 1.91 & 1.13 & 2.19 & 1.69 & 18.75 & 25.4 \\
& SeqKD                     & 36.01 & 13.20 & 29.27 & 33.85 & 2.03 & 1.29 & 6.35 & 2.48 & 29.21 & 39.6 \\
& KD                        & 44.81 & 17.80 & 29.27 & 38.91 & 3.10 & 1.41 & 7.04 & 3.48 & 35.13 & 47.6 \\
\rowcolor{gray!25}
& \shortopd\ (ours)         & \best{85.75} & \best{50.40} & \best{56.10} & \best{66.93} & \best{4.18} & \best{1.99} & \best{7.41} & \best{4.89} & \best{55.48} & \best{75.2} \\
\midrule
\multirow{6}{*}{\shortstack[l]{Qwen3-8B\\No-thinking}}
& Dense teacher             & 93.10 & 81.80 & 86.59 & 80.54 & 6.63 & 4.26 & 8.80 & 6.95 & 76.05 & 100.0 \\
& Pruned ($-25\%$)          & 0.83 & 1.80 & 0.00 & 0.00 & 1.03 & 1.01 & 1.00 & 1.01 & 5.39 & 7.1 \\
\cmidrule(lr){2-12}
& SFT w/o KD                & 39.58 & 10.80 & 15.24 & 21.01 & 2.16 & 1.51 & 3.10 & 2.03 & 21.83 & 28.7 \\
& SeqKD                     & 62.32 & 26.00 & 28.66 & 33.07 & 2.59 & 1.33 & 4.03 & 2.53 & 31.86 & 41.9 \\
& KD                        & 58.30 & 24.40 & 26.83 & 29.57 & 2.70 & 1.28 & 4.33 & 2.19 & 30.51 & 40.1 \\
\rowcolor{gray!25}
& \shortopd\ (ours)         & \best{75.74} & \best{40.80} & \best{40.24} & \best{42.80} & \best{3.38} & \best{1.45} & \best{6.19} & \best{3.29} & \best{42.84} & \best{56.3} \\
\midrule
\multirow{6}{*}{\shortstack[l]{Qwen3-8B\\Thinking}}
& Dense teacher             & 94.47 & 81.80 & 81.71 & 85.21 & 6.80 & 4.66 & 8.80 & 7.29 & 77.34 & 100.0 \\
& Pruned ($-25\%$)          & 0.15 & 2.00 & 0.00 & 0.00 & 1.01 & 1.00 & 1.00 & 1.01 & 5.30 & 6.9 \\
\cmidrule(lr){2-12}
& SFT w/o KD                & 61.33 & 23.80 & 24.39 & 29.18 & 3.33 & 1.66 & 6.70 & 2.81 & 35.46 & 45.8 \\
& SeqKD                     & 58.68 & 25.40 & 16.46 & 32.68 & 2.39 & 1.24 & 7.49 & 2.68 & 33.90 & 43.8 \\
& KD                        & 58.68 & 16.20 & 23.78 & 33.07 & 2.84 & 1.14 & 7.51 & 2.49 & 33.94 & 43.9 \\
\rowcolor{gray!25}
& \shortopd\ (ours)         & \best{87.26} & \best{47.40} & \best{49.39} & \best{55.64} & \best{4.73} & \best{1.89} & \best{7.98} & \best{4.41} & \best{53.71} & \best{69.4} \\
\bottomrule
\end{tabular}
}
\end{table}

\paragraph{Evaluation.}
We evaluate free-form generation only, on eight task families in the
same three domains:
\begin{itemize}
\item \textbf{Math}: GSM8K~\citep{cobbe2021gsm8k} and
MATH-500~\citep{hendrycks2021math}, scored by answer-match
accuracy. For 4B Instruct and 8B no-thinking, MATH-500 uses Qwen-style
long decoding (a $16$k generation cap) with the recommended sampling
parameters. The remaining no-thinking and Thinking protocol details are
given in Appendix~\ref{app:eval}.
\item \textbf{Code}: HumanEval~\citep{chen2021codex} and
MBPP~\citep{austin2021program}, scored by execution
\textsc{pass}@$1$.
\item \textbf{Open-ended}: Alpaca, QA, Summarization, and
MT-Bench~\citep{zheng2023mtbench}, judged on a $1$--$10$ scale by
GPT-5.5, an independent judge that is neither the teacher nor a Qwen
model.
\end{itemize}
Tables report
math/code scores as percentages and judge scores on the $1$--$10$
scale; the Avg column normalizes judge scores by $\times 10$ and
averages the eight task families.

\subsection{Main generation recovery}
\label{sec:main}

Table~\ref{tab:main-results} reports the main results. Structured
compression alone nearly destroys generation: the normalized average drops from
$75.17$ for the dense teacher to $5.71$ after pruning. One epoch of
SFT w/o KD recovers some ability ($21.19$ Avg), SeqKD improves on it
($28.60$), and KD adds a modest increase to $30.52$, but all
off-policy baselines remain far below on-policy distillation: \shortopd\
reaches $48.46$ Avg, nearly $9\times$ the untrained score and an
$18$-point gain over the best off-policy baseline.
The 4B Thinking backbone exhibits the same ordering: SFT w/o KD, SeqKD, and
KD reach $18.75$, $29.21$, and $35.13$ Avg, respectively, while
\shortopd\ reaches $55.48$, or $75.2\%$ of its dense teacher.

The scale-up results support the same conclusion. On 8B no-thinking,
depth pruning reduces Avg from $76.05$ to $5.39$; \shortopd\ restores it
to $42.84$ ($56.3\%$ of the teacher), $10.98$ points above the strongest
offline baseline (SeqKD at $31.86$). On 8B Thinking, depth pruning reduces Avg from $77.34$ to
$5.30$; \shortopd\ restores it to $53.71$ ($69.4\%$ of the teacher),
$18.25$ points above SFT w/o KD at $35.46$, and leads every generation task. Because the
available 8B no-thinking and Thinking evaluations use the distinct decoding
protocols documented in the caption, these absolute scores are used for
within-block recovery comparisons rather than a mode-to-mode ranking.

The gains are not confined to math: \shortopd\ improves code execution and
all four open-ended tasks. With no labels or verifiers, recovery comes solely
from matching the unpruned teacher on the student's own states.

\begin{figure}[t!]
\centering
\includegraphics[width=\linewidth]{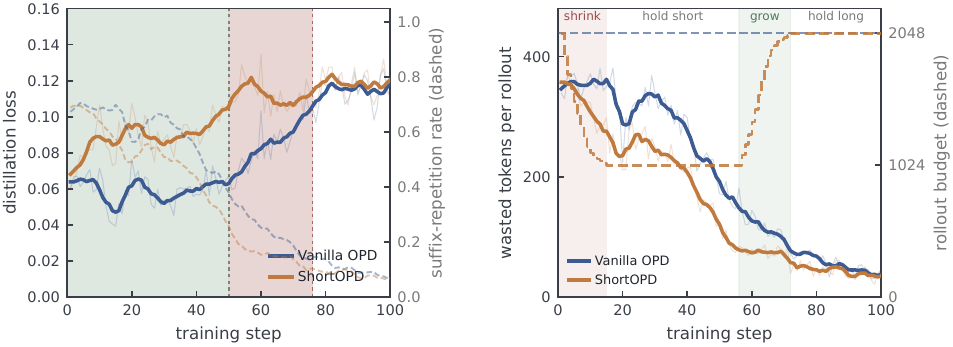}
\caption{Revisiting the Vanilla OPD warm-up of
Figure~\ref{fig:intro-prune-rep}(b).
\textbf{(a)} Distillation loss (solid, left axis) and suffix-repetition
rate (dashed, right axis) for Vanilla OPD and \shortopd. \shortopd's loss
reaches its plateau within roughly $40$--$50$ steps versus around $80$
for Vanilla: green shading marks \shortopd's warm-up, red the additional
Vanilla-only delay.
\textbf{(b)} Wasted (repetitive-suffix) tokens per rollout (solid, left
axis) and rollout budgets (dashed, right axis). \shortopd\ shrinks its
budget to $1024$ during the repetitive warm-up and restores it once clean
truncations dominate (shaded actions), removing the wasted mass sooner;
its effective length still catches Vanilla by step $100$.}
\label{fig:shortopd-traj}
\end{figure}

\subsection{\shortopd: closed-loop horizon control in practice}
\label{sec:shortopd-exp}

\paragraph{Closed-loop recovery trajectory.}
Figure~\ref{fig:shortopd-traj} shows that \shortopd\ shortens the
repetitive warm-up. With Vanilla fixed at $H{=}2048$, $55$--$75\%$ of
rollouts repeat and the loss reaches its post-warm-up plateau only after
roughly $80$ steps. \shortopd\ reaches this regime in about $40$--$50$
steps while reducing wasted suffix tokens. Its budget falls from $2048$
to $1024$ during high repetition and returns to $2048$ as clean
truncations take over. The controller therefore saves early rollout
compute without sacrificing long-generation capacity later in recovery.

\paragraph{Vanilla versus gated budgets.}
Figure~\ref{fig:shortopd-efficiency} compares quality and cost across
rollout schedules. Relative to Vanilla fixed at $2048$, \shortopd\ lowers
mean generation time over the first $100$ steps by $29\%$
($35.4\rightarrow25.1$s per step) and end-to-end wall-clock from $11.1$
to $8.5$ hours. Across fixed $2048$, \shortopd, and fixed $8192$, Avg
varies only from $48.5$ to $50.2$, whereas generated rollout tokens range
from $250$M to $869$M. \shortopd\ finishes in $8.5$ hours with $250$M
tokens, using $76\%$ less time than fixed $8192$ and $24\%$ less than
fixed $2048$ while remaining within $1.7$ Avg points of both. In contrast,
fixed $8192$ spends $869$M tokens and $35.9$ hours for only $+0.6$ Avg
over fixed $2048$.

\FloatBarrier

\subsection{Scaling}
\label{sec:scaling}

Table~\ref{tab:opd-scaling} tests whether the recovery recipe of
\shortopd\ continues to improve with additional exposure. Across one,
two, and three epochs on the compressed Qwen3-4B-Instruct, gains are
broad rather than concentrated in a single aggregate: GSM8K, MATH-500,
HumanEval, MBPP, Alpaca, QA, and MT-Bench all improve from the first to
the final checkpoint, with the overall Avg rising from $48.46$ to
$55.41$ ($73.7\%$ of the teacher). On-policy recovery is therefore
scalable: more rollout exposure keeps converting into generation
quality, and pushing this scaling further is a natural direction for
future work.

\begin{table}[ht]
\centering
\small
\renewcommand{\arraystretch}{1.18}
\setlength{\tabcolsep}{1.8pt}
\caption{Per-domain epoch scaling for \shortopd\ on the compressed
Qwen3-4B-Instruct. Math/code are percentages; open-ended tasks are judge
scores on a $1$--$10$ scale. Avg normalizes judge scores by $\times 10$,
and T\% is relative to the Instruct teacher Avg.}
\label{tab:opd-scaling}
\begin{tabular}{lcccccccccc}
\toprule
 & \multicolumn{2}{c}{Math} & \multicolumn{2}{c}{Code}
 & \multicolumn{4}{c}{Open-ended} & \multicolumn{2}{c}{Overall} \\
\cmidrule(lr){2-3}\cmidrule(lr){4-5}\cmidrule(lr){6-9}\cmidrule(lr){10-11}
Checkpoint & GSM8K & MATH & HumanEval & MBPP & Alpaca & QA & Sum & MT-B & Avg & T\% \\
\midrule
1 epoch / step 710   & 62.70 & 42.00 & 43.90 & 49.81 & 4.68 & 2.04 & 7.94 & 4.28 & 48.46 & 64.5 \\
2 epochs / step 1420 & 70.66 & 50.60 & 51.83 & 56.42 & 4.98 & 2.20 & \best{8.16} & 4.47 & 53.45 & 71.1 \\
3 epochs / step 2127 & \best{72.71} & \best{55.20} & \best{54.27} & \best{57.20} & \best{5.14} & \best{2.21} & 8.15 & \best{4.89} & \best{55.41} & \best{73.7} \\
\bottomrule
\end{tabular}
\end{table}

\subsection{Signal controls: dense versus sparse and off-policy}
\label{sec:rl-opd}

The \shortopd-versus-SeqKD and \shortopd-versus-KD contrasts in
Table~\ref{tab:main-results} test whether teacher information is sufficient
without on-policy states: with the same frozen teacher, moving its signal
off-policy costs $19.9$ Avg points (SeqKD), and dense teacher-forced
distillation does not close the gap (KD trails \shortopd\ by $17.9$
points).

To isolate the supervision axis, we run a matched Math+Code comparison
between sparse-reward RLVR and \shortopd, instantiating RLVR with both PPO
and GRPO. For each model size, all methods start from the same
$25\%$ depth-pruned initialization and train on the same shuffled
$29{,}585$-prompt corpus: $7{,}473$ GSM8K, $7{,}500$ MATH, $14{,}139$
verified OpenCodeInstruct, and $473$ MBPP train-side prompts.

\begin{figure}[t!]
\centering
% The source panels have equal canvases but different axis widths because the
% bar chart needs a wider y-label margin.  Use asymmetric containers and a
% fixed common height so the two data rectangles render at the same size.
\begin{subfigure}[b]{0.475\linewidth}
\centering
\includegraphics[width=\linewidth]{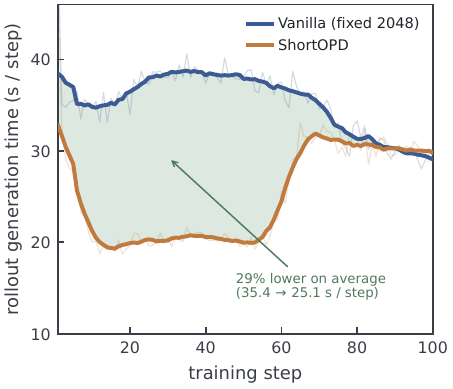}
\caption{Per-step rollout generation time}
\end{subfigure}\hfill%
\begin{subfigure}[b]{0.522\linewidth}
\centering
\includegraphics[width=\linewidth,height=0.781\linewidth]{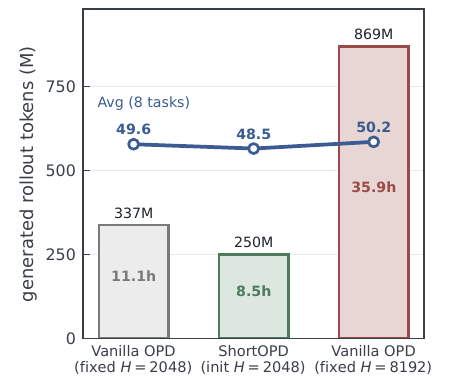}
\caption{Quality vs.\ rollout cost across schedules}
\end{subfigure}
\caption{What uncontrolled rollout budgets cost. \textbf{(a)} Per-step
rollout-generation time for Vanilla (fixed 2048) vs.\ \shortopd\ (faint: raw;
solid: moving average) over the same first $100$ steps as
Figure~\ref{fig:shortopd-traj}; the shaded gap is saved warm-up compute
($35.4\rightarrow25.1$s per step on average). \textbf{(b)} Across three schedules,
the eight-task Avg (blue line) varies by less than $2$ points while the
generated rollout tokens (bars; \shortopd\ in green, Vanilla fixed 8192 in red)
vary by more than $3\times$. Wall-clock is annotated inside each bar.}
\label{fig:shortopd-efficiency}
\end{figure}

The code rows
contain one hidden executable test, used only by PPO and GRPO; \shortopd\
uses the same prompts but ignores rewards and tests, receiving dense logits
from the frozen teacher on its own rollouts. All runs use batch $64$, eight
rollouts per prompt, the no-thinking template, and the step-$462$ checkpoint
at data exhaustion. Table~\ref{tab:rlvr-opd} reports the unified full-set
evaluation.

At 4B, \shortopd\ reaches $58.68$/$38.00$ on GSM8K/MATH-500 and
$41.46$/$44.36$ on HumanEval/MBPP. PPO scores zero on all four, while GRPO
reaches only $0.91$/$0.80$ on math and zero on code. Scaling to 8B raises
PPO's GSM8K score to $14.78$, but leaves MATH-500 at $2.40$ and both code
benchmarks at zero; \shortopd\ instead reaches $76.88$/$39.60$ on math and
$40.24$/$42.41$ on code. Across the full generation suite, \shortopd\
achieves $38.85$ and $41.62$ Avg at 4B and 8B, versus at most $5.38$ and
$7.68$ for sparse-reward RLVR. The held-out MC averages show the same
separation: $61.63$/$75.29$ for \shortopd, $48.17$/$62.30$ for PPO, and
$41.27$/$59.11$ for GRPO. Thus model scale alone does not repair the
sparse-signal bottleneck. When the compressed model rarely samples a
correct answer, a sparse reward has few successful trajectories to promote,
whereas the dense teacher supplies a token-level correction on every
rollout regardless of final-answer correctness.

\begin{table}[h]
\centering
\small
\renewcommand{\arraystretch}{1.16}
\setlength{\tabcolsep}{5.2pt}
\caption{Dense versus sparse on-policy supervision on the matched
Math+Code recovery corpus. Math/code scores are accuracy or
\textsc{pass}@$1$ (\%). Gen. Avg is the normalized average over all eight
generation tasks; MATH-500 uses the $16$k official-style protocol. Dense
and pruned rows are carried over from Table~\ref{tab:main-results}.
Bold marks the best recovery method for each model size.}
\label{tab:rlvr-opd}
\begin{tabular}{llccccc}
\toprule
Model & Method & GSM8K & MATH-500 & HumanEval & MBPP & Gen. Avg \\
\midrule
\multirow{5}{*}{\shortstack[l]{Qwen3-4B\\Instruct}}
& Teacher & 88.10 & 89.40 & 75.00 & 80.54 & 75.17 \\
& Pruned & 1.14 & 1.60 & 0.00 & 0.00 & 5.71 \\
\cmidrule(lr){2-7}
& PPO & 0.00 & 0.00 & 0.00 & 0.00 & 5.09 \\
& GRPO & 0.91 & 0.80 & 0.00 & 0.00 & 5.38 \\
\rowcolor{gray!25}
& \shortopd\ (ours) & \best{58.68} & \best{38.00} & \best{41.46} & \best{44.36} & \best{38.85} \\
\midrule
\multirow{5}{*}{\shortstack[l]{Qwen3-8B\\No-thinking}}
& Teacher & 93.10 & 81.80 & 86.59 & 80.54 & 76.05 \\
& Pruned & 0.83 & 1.80 & 0.00 & 0.00 & 5.39 \\
\cmidrule(lr){2-7}
& PPO & 14.78 & 2.40 & 0.00 & 0.00 & 7.68 \\
& GRPO & 0.99 & 2.20 & 0.00 & 0.00 & 5.53 \\
\rowcolor{gray!25}
& \shortopd\ (ours) & \best{76.88} & \best{39.60} & \best{40.24} & \best{42.41} & \best{41.62} \\
\bottomrule
\end{tabular}
\end{table}

\subsection{Recovery-corpus domain ablation}
\label{sec:domain}

Because \shortopd\ is reward-free, the recovery corpus is only a
prompt distribution; we test how much its composition matters.
Table~\ref{tab:domain-ablation} recovers the same compressed student
with leave-one-domain-out recovery corpora. To keep the comparison matched,
each ablated corpus keeps the same number of prompts and the same $710$
optimizer steps as the full mixture by uniformly resampling from the two
remaining domains; only the removed domain changes. This isolates prompt
composition from training exposure.

\begin{table}[H]
\centering
\small
\renewcommand{\arraystretch}{1.22}
\setlength{\tabcolsep}{4.5pt}
\caption{Leave-one-domain-out recovery-corpus ablation for \shortopd\
on the compressed Qwen3-4B-Instruct. Math/code are percentages; MATH
uses the official MATH-500 protocol. Open-ended tasks are judge scores on a
$1$--$10$ scale; Avg normalizes them by $\times 10$.
\textcolor{red!70!black}{Red} marks the domain most directly damaged by
the removed slice; the shaded row is the full-mixture reference.}
\label{tab:domain-ablation}
\begin{tabular}{lccccccccc}
\toprule
 & \multicolumn{2}{c}{Math} & \multicolumn{2}{c}{Code}
 & \multicolumn{4}{c}{Open-ended} & \multicolumn{1}{c}{Overall} \\
\cmidrule(lr){2-3}\cmidrule(lr){4-5}\cmidrule(lr){6-9}\cmidrule(lr){10-10}
Recovery corpus & GSM8K & MATH & HumanEval & MBPP & Alpaca & QA & Sum & MT-B & Avg \\
\midrule
\rowcolor{gray!12}
Full mixture       & 62.70 & 42.00 & 43.90 & 49.81 & 4.68 & 2.04 & 7.94 & 4.28 & 48.46 \\
w/o code           & 58.45 & 33.40 & \textcolor{red!70!black}{0.61}  & \textcolor{red!70!black}{2.33}  & 4.47 & 2.15 & 7.88 & 3.99 & 34.96 \\
w/o math           & \textcolor{red!70!black}{8.04}  & \textcolor{red!70!black}{6.00}  & 28.66 & 43.19 & 3.96 & 1.73 & 7.19 & 2.94 & 30.51 \\
w/o general        & 54.21 & 28.80 & 48.17 & 50.19 & \textcolor{red!70!black}{3.01} & \textcolor{red!70!black}{1.51} & \textcolor{red!70!black}{5.41} & \textcolor{red!70!black}{3.15} & 39.02 \\
\bottomrule
\end{tabular}
\end{table}

The resulting pattern is direct. Removing math nearly eliminates mathematical
recovery, dropping GSM8K from $62.70$ to $8.04$ and official MATH-500 from
$42.00$ to $6.00$. Removing code also lowers math, but collapses execution
benchmarks (HumanEval $0.61$, MBPP $2.33$), showing that code rollouts provide
a highly domain-specific repair signal. Removing general instruction data
reduces both math and open-ended quality in this run, despite improving the
narrow code scores. Thus \shortopd's dense token-level signal can transfer across
related states, but the on-policy state distribution still needs coverage
of the capabilities we want to recover.

\subsection{Extension to width and hybrid pruning}
\label{sec:structure-extension}

The main results remove transformer layers. We next extend the same recovery
comparison from depth pruning to width-only and hybrid width--depth
configurations, following the structured-pruning configurations used by
Minitron~\citep{muralidharan2024minitron,sreenivas2024minitron}.
Table~\ref{tab:structure-extension} compares KD and \shortopd\ from the
identical pruned initialization under the same $710$-step prompt budget.
Across all four scale--structure pairs, \shortopd\ improves generation Avg by
$6.44$--$9.01$ points over KD. Before recovery, pruning reduces generation
Avg to $16.17$--$19.71$; \shortopd\ restores it to $51.46$--$58.86$, or
$69.0$--$78.2\%$ of the corresponding dense teacher. Thus the recovery gain
is not specific to deleting depth: it transfers to compression along width
and jointly along width and depth. Complementary multiple-choice results are
reported in Appendix~\ref{app:mc}.

\begin{table}[ht]
\centering
\small
\renewcommand{\arraystretch}{0.95}
\setlength{\tabcolsep}{4.4pt}
\caption{Generation recovery under width-only and hybrid pruning. Parenthetical
percentages report the actual parameter reduction for each configuration
(Appendix~\ref{tab:structure-config}); bold marks the best recovery result
within each pruned configuration.}
\label{tab:structure-extension}
\resizebox{\linewidth}{!}{%
\begin{tabular}{llccccccccc}
\toprule
& & \multicolumn{2}{c}{Math} & \multicolumn{2}{c}{Code} & \multicolumn{4}{c}{Open-ended} & \multicolumn{1}{c}{Overall} \\
\cmidrule(lr){3-4}\cmidrule(lr){5-6}\cmidrule(lr){7-10}\cmidrule(lr){11-11}
Structure & Recovery & GSM8K & MATH & HumanEval & MBPP & Alpaca & QA & Sum & MT-B & Gen. Avg \\
\midrule
\rowcolor{gray!10}
\multicolumn{11}{c}{\textit{Qwen3-4B-Instruct}} \\
Dense & Teacher & 88.02 & 87.00 & 75.61 & 80.16 & 6.56 & 4.34 & 8.76 & 6.92 & 74.58 \\
\cmidrule(lr){1-11}
\multirow{3}{*}{Width (29.5\%)} & Pruned & 2.81 & 5.80 & 1.22 & 0.00 & 2.52 & 1.30 & 6.95 & 2.02 & 17.23 \\
& KD & 47.76 & 22.80 & 32.93 & 45.14 & 5.31 & 2.42 & 8.20 & 4.16 & 43.70 \\
\rowcolor{gray!25}
& \shortopd\ & \best{69.75} & \best{46.20} & \best{35.37} & \best{49.03} & \best{5.71} & \best{2.44} & \best{8.31} & \best{4.67} & \best{51.46} \\
\cmidrule(lr){1-11}
\multirow{3}{*}{Hybrid (29.4\%)} & Pruned & 14.10 & 6.20 & 2.44 & 7.39 & 2.75 & 1.27 & 6.69 & 2.04 & 19.71 \\
& KD & 55.19 & 30.00 & \best{43.29} & 47.86 & 4.99 & 2.15 & 8.00 & 4.15 & 46.15 \\
\rowcolor{gray!25}
& \shortopd\ & \best{73.92} & \best{54.40} & 42.68 & \best{56.81} & \best{5.60} & \best{2.46} & \best{8.21} & \best{5.08} & \best{55.16} \\
\midrule
\rowcolor{gray!10}
\multicolumn{11}{c}{\textit{Qwen3-8B No-thinking}} \\
Dense & Teacher & 92.19 & 79.20 & 83.54 & 78.60 & 6.44 & 4.36 & 8.90 & 7.14 & 75.24 \\
\cmidrule(lr){1-11}
\multirow{3}{*}{Width (21.0\%)} & Pruned & 9.48 & 5.40 & 4.88 & 15.56 & 2.30 & 1.23 & 6.83 & 1.86 & 19.67 \\
& KD & 67.10 & 35.40 & 50.61 & 49.03 & 5.62 & 2.58 & 8.38 & 5.14 & 52.42 \\
\rowcolor{gray!25}
& \shortopd\ & \best{78.85} & \best{50.60} & \best{54.88} & \best{56.03} & \best{5.96} & \best{2.96} & \best{8.50} & \best{5.63} & \best{58.86} \\
\cmidrule(lr){1-11}
\multirow{3}{*}{Hybrid (21.2\%)} & Pruned & 4.93 & 6.80 & 10.98 & 10.12 & 2.25 & 1.18 & 4.25 & 1.98 & 16.17 \\
& KD & 59.51 & 30.80 & 50.00 & 53.31 & 4.80 & 1.84 & 8.50 & 4.42 & 48.66 \\
\rowcolor{gray!25}
& \shortopd\ & \best{76.72} & \best{47.40} & \best{57.93} & \best{56.03} & \best{5.45} & \best{2.19} & \best{8.55} & \best{5.33} & \best{56.65} \\
\bottomrule
\end{tabular}
}
\end{table}

\section{Conclusion}
\label{sec:conclusion}
Structured pruning can leave latent capability in the model's search space
while failing to promote it during generation, making recovery a distributional
repair problem. Across Qwen3 at 4B and 8B, in no-thinking and Thinking modes,
and after depth, width, or hybrid pruning, on-policy self-distillation from the
pre-compression model substantially outperforms SFT w/o KD, SeqKD, and KD;
in matched Math+Code controls it also outperforms sparse-reward RLVR. Its main
cost is low-quality early rollouts: long horizons waste tokens on repetitive
suffixes before normal loss descent. \shortopd's repetition-gated,
truncation-aware short-to-long budget matches fixed short and long horizons
within two points while using fewer rollout tokens and focusing early teacher
compute on higher-signal prefixes. We therefore recommend short-to-long
on-policy distillation as a default recovery step after structured pruning.

\section{Limitations}
\label{sec:limitations}
We study BI depth, width-only, and hybrid pruning on Qwen3 models at 4B and
8B. Broader model families, compression ratios, and scales are needed to
establish generality and the point at which compression becomes too aggressive
for lightweight recovery.

\clearpage
\bibliographystyle{unsrtnat}
\bibliography{main}

@inproceedings{agarwal2024gkd,
  title={On-Policy Distillation of Language Models: Learning from Self-Generated Mistakes},
  author={Agarwal, Rishabh and Vieillard, Nino and Zhou, Yongchao and Stanczyk, Piotr and Garea, Sabela Ramos and Geist, Matthieu and Bachem, Olivier},
  booktitle={International Conference on Learning Representations (ICLR)},
  year={2024}
}

@inproceedings{an2024flap,
  title={Fluctuation-Based Adaptive Structured Pruning for Large Language Models},
  author={An, Yongqi and Zhao, Xu and Yu, Tao and Tang, Ming and Wang, Jinqiao},
  booktitle={AAAI Conference on Artificial Intelligence (AAAI)},
  year={2024}
}

@inproceedings{ashkboos2024slicegpt,
  title={SliceGPT: Compress Large Language Models by Deleting Rows and Columns},
  author={Ashkboos, Saleh and Croci, Maximilian L and Nascimento, Marcelo Gennari do and Hoefler, Torsten and Hensman, James},
  booktitle={International Conference on Learning Representations (ICLR)},
  year={2024}
}

@article{austin2021program,
  title={Program Synthesis with Large Language Models},
  author={Austin, Jacob and Odena, Augustus and Nye, Maxwell and Bosma, Maarten and Michalewski, Henryk and Dohan, David and Jiang, Ellen and Cai, Carrie and Terry, Michael and Le, Quoc and Sutton, Charles},
  journal={arXiv preprint arXiv:2108.07732},
  year={2021}
}

@article{chen2021codex,
  title={Evaluating Large Language Models Trained on Code},
  author={Chen, Mark and Tworek, Jerry and Jun, Heewoo and Yuan, Qiming and Pinto, Henrique Ponde de Oliveira and Kaplan, Jared and Edwards, Harri and Burda, Yuri and Joseph, Nicholas and Brockman, Greg and others},
  journal={arXiv preprint arXiv:2107.03374},
  year={2021}
}

@article{cobbe2021gsm8k,
  title={Training Verifiers to Solve Math Word Problems},
  author={Cobbe, Karl and Kosaraju, Vineet and Bavarian, Mohammad and others},
  journal={arXiv preprint arXiv:2110.14168},
  year={2021}
}

@article{deepseekr1,
  title={DeepSeek-R1: Incentivizing Reasoning Capability in LLMs via Reinforcement Learning},
  author={{DeepSeek-AI}},
  journal={arXiv preprint arXiv:2501.12948},
  year={2025}
}

@article{dery2024everybody,
  title={Everybody Prune Now: Structured Pruning of LLMs with Only Forward Passes},
  author={Dery, Lucio and Kolawole, Steven and Kagey, Jean-Fran{\c{c}}ois and Smith, Virginia and Neubig, Graham and Talwalkar, Ameet},
  journal={arXiv preprint arXiv:2402.05406},
  year={2024}
}

@article{duan2026circular,
  title={Circular Reasoning: Understanding Self-Reinforcing Loops in Large Reasoning Models},
  author={Duan, Zenghao and Pang, Liang and Wei, Zihao and Duan, Wenbin and Tian, Yuxin and Xu, Shicheng and Deng, Jingcheng and Yin, Zhiyi and Cheng, Xueqi},
  journal={arXiv preprint arXiv:2601.05693},
  year={2026}
}

@inproceedings{frantar2023sparsegpt,
  title={SparseGPT: Massive Language Models Can Be Accurately Pruned in One-Shot},
  author={Frantar, Elias and Alistarh, Dan},
  booktitle={International Conference on Machine Learning (ICML)},
  year={2023}
}

@article{grattafiori2024llama3,
  title={The Llama 3 Herd of Models},
  author={Grattafiori, Aaron and Dubey, Abhimanyu and Jauhri, Abhinav and others},
  journal={arXiv preprint arXiv:2407.21783},
  year={2024}
}

@inproceedings{gromov2025unreasonable,
  title={The Unreasonable Ineffectiveness of the Deeper Layers},
  author={Gromov, Andrey and Tirumala, Kushal and Shapourian, Hassan and Glorioso, Paolo and Roberts, Daniel A},
  booktitle={International Conference on Learning Representations (ICLR)},
  year={2025}
}

@inproceedings{gu2024minillm,
  title={MiniLLM: On-Policy Distillation of Large Language Models},
  author={Gu, Yuxian and Dong, Li and Wei, Furu and Huang, Minlie},
  booktitle={International Conference on Learning Representations (ICLR)},
  year={2024}
}

@inproceedings{hendrycks2021math,
  title={Measuring Mathematical Problem Solving With the MATH Dataset},
  author={Hendrycks, Dan and Burns, Collin and Kadavath, Saurav and Arora, Akul and Basart, Steven and Tang, Eric and Song, Dawn and Steinhardt, Jacob},
  booktitle={NeurIPS Datasets and Benchmarks},
  year={2021}
}

@inproceedings{hendrycks2021mmlu,
  title={Measuring Massive Multitask Language Understanding},
  author={Hendrycks, Dan and Burns, Collin and Basart, Steven and Zou, Andy and Mazeika, Mantas and Song, Dawn and Steinhardt, Jacob},
  booktitle={International Conference on Learning Representations (ICLR)},
  year={2021}
}

@inproceedings{he2026demystifying,
  title={Demystifying When Pruning Works via Representation Hierarchies},
  author={He, Shwai and Sun, Guoheng and Zhang, Haichao and Fu, Yun and Li, Ang},
  booktitle={International Conference on Machine Learning (ICML)},
  year={2026}
}

@article{hinton2015distilling,
  title={Distilling the Knowledge in a Neural Network},
  author={Hinton, Geoffrey and Vinyals, Oriol and Dean, Jeff},
  journal={arXiv preprint arXiv:1503.02531},
  year={2015}
}

@inproceedings{hiraoka2025repetition,
  title={Repetition Neurons: How Do Language Models Produce Repetitions?},
  author={Hiraoka, Tatsuya and Inui, Kentaro},
  booktitle={Proceedings of the 2025 Conference of the Nations of the Americas Chapter of the Association for Computational Linguistics: Human Language Technologies (Volume 2: Short Papers)},
  pages={483--495},
  doi={10.18653/v1/2025.naacl-short.41},
  year={2025}
}

@inproceedings{holtzman2020degeneration,
  title     = {The Curious Case of Neural Text Degeneration},
  author    = {Holtzman, Ari and Buys, Jan and Du, Li and Forbes, Maxwell and Choi, Yejin},
  booktitle = {International Conference on Learning Representations (ICLR)},
  year      = {2020}
}

@inproceedings{hu2026logits,
  title={Logits-Based Block Pruning with Affine Transformations for Large Language Models},
  author={Hu, Zekun and Xu, Yichu and Zhan, De-Chuan},
  booktitle={Findings of the Association for Computational Linguistics: EACL 2026},
  pages={3722--3736},
  doi={10.18653/v1/2026.findings-eacl.193},
  year={2026}
}

@inproceedings{kim2016sequence,
  title={Sequence-Level Knowledge Distillation},
  author={Kim, Yoon and Rush, Alexander M},
  booktitle={Empirical Methods in Natural Language Processing (EMNLP)},
  year={2016}
}

@article{kim2024shortened,
  title={Shortened LLaMA: Depth Pruning for Large Language Models with Comparison of Retraining Methods},
  author={Kim, Bo-Kyeong and Kim, Geonmin and Kim, Tae-Ho and Castells, Thibault and Choi, Shinkook and Shin, Junho and Song, Hyoung-Kyu},
  journal={arXiv preprint arXiv:2402.02834},
  year={2024}
}

@inproceedings{ko2024distillm,
  title={DistiLLM: Towards Streamlined Distillation for Large Language Models},
  author={Ko, Jongwoo and Kim, Sungnyun and Chen, Tianyi and Yun, Se-Young},
  booktitle={International Conference on Machine Learning (ICML)},
  year={2024}
}

@inproceedings{lambert2025tulu3,
  title={Tulu 3: Pushing Frontiers in Open Language Model Post-Training},
  author={Lambert, Nathan and Morrison, Jacob and Pyatkin, Valentina and Huang, Shengyi and Ivison, Hamish and Brahman, Faeze and others},
  booktitle={Conference on Language Modeling (COLM)},
  year={2025}
}

@inproceedings{lin2020imitkd,
  title={Autoregressive Knowledge Distillation through Imitation Learning},
  author={Lin, Alexander and Wohlwend, Jeremy and Chen, Howard and Lei, Tao},
  booktitle={Empirical Methods in Natural Language Processing (EMNLP)},
  year={2020}
}

@inproceedings{lin2024awq,
  title={AWQ: Activation-aware Weight Quantization for LLM Compression and Acceleration},
  author={Lin, Ji and Tang, Jiaming and Tang, Haotian and Yang, Shang and Chen, Wei-Ming and Wang, Wei-Chen and Xiao, Guangxuan and Dang, Xingyu and Gan, Chuang and Han, Song},
  booktitle={Proceedings of Machine Learning and Systems (MLSys)},
  year={2024}
}

@inproceedings{ma2023llmpruner,
  title={LLM-Pruner: On the Structural Pruning of Large Language Models},
  author={Ma, Xinyin and Fang, Gongfan and Wang, Xinchao},
  booktitle={Advances in Neural Information Processing Systems (NeurIPS)},
  year={2023}
}

@inproceedings{men2025shortgpt,
  title={ShortGPT: Layers in Large Language Models are More Redundant Than You Expect},
  author={Men, Xin and Xu, Mingyu and Zhang, Qingyu and Yuan, Qianhao and Wang, Bingning and Lin, Hongyu and Lu, Yaojie and Han, Xianpei and Chen, Weipeng},
  booktitle={Findings of the Association for Computational Linguistics: ACL 2025},
  year={2025}
}

@inproceedings{muralidharan2024minitron,
  title={Compact Language Models via Pruning and Knowledge Distillation},
  author={Muralidharan, Saurav and Sreenivas, Sharath Turuvekere and Joshi, Raviraj and Chochowski, Marcin and Patwary, Mostofa and Shoeybi, Mohammad and Catanzaro, Bryan and Kautz, Jan and Molchanov, Pavlo},
  booktitle={Advances in Neural Information Processing Systems (NeurIPS)},
  year={2024}
}

@article{qwen3,
  title={Qwen3 Technical Report},
  author={{Qwen Team}},
  journal={arXiv preprint arXiv:2505.09388},
  year={2025}
}

@inproceedings{pipis2026wait,
  title={Wait, Wait, Wait... Why Do Reasoning Models Loop?},
  author={Pipis, Charilaos and Garg, Shivam and Kontonis, Vasilis and Shrivastava, Vaishnavi and Krishnamurthy, Akshay and Papailiopoulos, Dimitris},
  booktitle={International Conference on Machine Learning (ICML)},
  year={2026}
}

@inproceedings{ranzato2016sequence,
  title={Sequence Level Training with Recurrent Neural Networks},
  author={Ranzato, Marc'Aurelio and Chopra, Sumit and Auli, Michael and Zaremba, Wojciech},
  booktitle={International Conference on Learning Representations (ICLR)},
  year={2016}
}

@article{shao2024deepseekmath,
  title={DeepSeekMath: Pushing the Limits of Mathematical Reasoning in Open Language Models},
  author={Shao, Zhihong and Wang, Peiyi and Zhu, Qihao and Xu, Runxin and Song, Junxiao and others},
  journal={arXiv preprint arXiv:2402.03300},
  year={2024}
}

@article{shrestha2026limits,
  title={On the Limits of Layer Pruning for Generative Reasoning in Large Language Models},
  author={Shrestha, Safal and Shrestha, Anubhav and Nepal, Aadim and Kim, Minwu and Ross, Keith},
  journal={arXiv preprint arXiv:2602.01997},
  year={2026}
}

@article{sreenivas2024minitron,
  title={LLM Pruning and Distillation in Practice: The Minitron Approach},
  author={Sreenivas, Sharath Turuvekere and Muralidharan, Saurav and Joshi, Raviraj and Chochowski, Marcin and Mahabaleshwarkar, Ameya Sunil and Shen, Gerald and others},
  journal={arXiv preprint arXiv:2408.11796},
  year={2024}
}

@inproceedings{sun2024wanda,
  title={A Simple and Effective Pruning Approach for Large Language Models},
  author={Sun, Mingjie and Liu, Zhuang and Bair, Anna and Kolter, J Zico},
  booktitle={International Conference on Learning Representations (ICLR)},
  year={2024}
}

@inproceedings{thangarasa2025selfdata,
  title={Self-Data Distillation for Recovering Quality in Pruned Large Language Models},
  author={Thangarasa, Vithursan and Venkatesh, Ganesh and Lasby, Mike and Sinnadurai, Nish and Lie, Sean},
  booktitle={Proceedings of Machine Learning and Systems (MLSys)},
  volume={7},
  url={https://proceedings.mlsys.org/paper_files/paper/2025/file/af2d9fb5bcee19ef2dfa70d843520c97-Paper-Conference.pdf},
  year={2025}
}

@inproceedings{wen2023fdistill,
  title={f-Divergence Minimization for Sequence-Level Knowledge Distillation},
  author={Wen, Yuqiao and Li, Zichao and Du, Wenyu and Mou, Lili},
  booktitle={Annual Meeting of the Association for Computational Linguistics (ACL)},
  year={2023}
}

@article{wang2025fewer,
  title={When Fewer Layers Break More Chains: Layer Pruning Harms Test-Time Scaling in {LLM}s},
  author={Wang, Keyu and Lyu, Tian and Su, Guinan and Yin, Lu and Canini, Marco and Geiping, Jonas and Liu, Shiwei},
  journal={arXiv preprint arXiv:2510.22228},
  year={2025}
}

@article{wen2026illusion,
  title={The Benchmark Illusion: Pruned LLMs Can Pass Multiple Choice but Fail to Answer},
  author={Wen, Rui and Sun, Lu and Liu, Jiayang and Xu, Zesheng and Cong, Tianshuo and Li, Zheng},
  journal={arXiv preprint arXiv:2606.17609},
  year={2026}
}

@inproceedings{xia2024sheared,
  title={Sheared LLaMA: Accelerating Language Model Pre-training via Structured Pruning},
  author={Xia, Mengzhou and Gao, Tianyu and Zeng, Zhiyuan and Chen, Danqi},
  booktitle={International Conference on Learning Representations (ICLR)},
  year={2024}
}

@inproceedings{xu2022ditto,
  title     = {Learning to Break the Loop: Analyzing and Mitigating Repetitions for Neural Text Generation},
  author    = {Xu, Jin and Liu, Xiaojiang and Yan, Jianhao and Cai, Deng and Li, Huayang and Li, Jian},
  booktitle = {Advances in Neural Information Processing Systems (NeurIPS)},
  year      = {2022}
}

@inproceedings{yang2024laco,
  title={LaCo: Large Language Model Pruning via Layer Collapse},
  author={Yang, Yifei and Cao, Zouying and Zhao, Hai},
  booktitle={Findings of the Association for Computational Linguistics: EMNLP 2024},
  year={2024}
}

@inproceedings{yu2025dapo,
  title={DAPO: An Open-Source LLM Reinforcement Learning System at Scale},
  author={Yu, Qiying and Zhang, Zheng and Zhu, Ruofei and Yuan, Yufeng and Zuo, Xiaochen and Yue, Yu and others},
  booktitle={Advances in Neural Information Processing Systems (NeurIPS)},
  year={2025}
}

@article{yuan2026vision,
  title={Vision-OPD: Learning to See Fine Details for Multimodal LLMs via On-Policy Self-Distillation},
  author={Yuan, Qianhao and Lou, Jie and Yu, Xing and Lin, Hongyu and Sun, Le and Han, Xianpei and Lu, Yaojie},
  journal={arXiv preprint arXiv:2605.18740},
  year={2026}
}

@inproceedings{zellers2019hellaswag,
  title={HellaSwag: Can a Machine Really Finish Your Sentence?},
  author={Zellers, Rowan and Holtzman, Ari and Bisk, Yonatan and Farhadi, Ali and Choi, Yejin},
  booktitle={Annual Meeting of the Association for Computational Linguistics (ACL)},
  year={2019}
}

@inproceedings{zheng2023mtbench,
  title={Judging LLM-as-a-Judge with MT-Bench and Chatbot Arena},
  author={Zheng, Lianmin and Chiang, Wei-Lin and Sheng, Ying and Zhuang, Siyuan and Wu, Zhanghao and Zhuang, Yonghao and Lin, Zi and Li, Zhuohan and Li, Dacheng and Xing, Eric P. and Zhang, Hao and Gonzalez, Joseph E. and Stoica, Ion},
  booktitle={Advances in Neural Information Processing Systems (NeurIPS) Datasets and Benchmarks Track},
  year={2023}
}

@inproceedings{yuan2025shortv,
  title={Shortv: Efficient multimodal large language models by freezing visual tokens in ineffective layers},
  author={Yuan, Qianhao and Zhang, Qingyu and Liu, Yanjiang and Chen, Jiawei and Lu, Yaojie and Lin, Hongyu and Zheng, Jia and Han, Xianpei and Sun, Le},
  booktitle={Proceedings of the IEEE/CVF International Conference on Computer Vision},
  pages={329--339},
  year={2025}
}

@article{schulman2017ppo,
  title={Proximal policy optimization algorithms},
  author={Schulman, John and Wolski, Filip and Dhariwal, Prafulla and Radford, Alec and Klimov, Oleg},
  journal={arXiv preprint arXiv:1707.06347},
  year={2017}
}

\clearpage

\beginappendix

\section{Training details}
\label{app:train}

\paragraph{Compression.}
Qwen3-4B-Instruct-2507 has
$36$ transformer blocks. BI is calibrated on a held-out corpus, and
the nine lowest-BI blocks are removed
(Table~\ref{tab:pruned-layers}); indices are reported before the
retained blocks are renumbered at export.
The Qwen3-4B-Thinking and Qwen3-8B experiments use the same $36\!\to\!27$
depth reduction, with BI calibrated separately for each backbone.
The frozen original model serves as the teacher and the compressed
student initializes the actor.

\begin{table}[h!]
\centering
\small
\renewcommand{\arraystretch}{1.12}
\setlength{\tabcolsep}{6pt}
\caption{Pruned configuration.}
\label{tab:pruned-layers}
\begin{tabular}{@{}cccc@{}}
\toprule
Original depth & Removed block IDs (zero-based) & Student depth & Param.\ reduction \\
\midrule
$36$ & $25$--$33$ (layers $26$--$34$, one-based) & $27$ & ${\sim}25\%$ \\
\bottomrule
\end{tabular}
\end{table}

For the width/hybrid structural-pruning comparison, importance scores are
calibrated on held-out text and shared by the two exports of each teacher,
following Minitron~\citep{muralidharan2024minitron,sreenivas2024minitron}.
Table~\ref{tab:structure-config} gives the resulting
architectures and exact unique-weight parameter counts. Attention head
dimensions and the $32$ query/$8$ key-value head layout are retained.

\begin{table}[h!]
\centering
\small
\renewcommand{\arraystretch}{1.12}
\setlength{\tabcolsep}{4.5pt}
\caption{Width-only and hybrid structural-pruning configurations. $L$, $H$, and
$I$ denote layer count, hidden size, and MLP intermediate size.}
\label{tab:structure-config}
\begin{tabular}{llrrrrr}
\toprule
Model & Structure & $L$ & $H$ & $I$ & Params & Reduction \\
\midrule
\multirow{2}{*}{\shortstack[l]{Qwen3-4B\\Instruct}}
& Width  & 36 & 2304 & 7680  & 3.111B & 29.5\% \\
& Hybrid & 33 & 2304 & 8704  & 3.114B & 29.4\% \\
\midrule
\multirow{2}{*}{\shortstack[l]{Qwen3-8B\\No-thinking}}
& Width  & 36 & 3584 & 10496 & 6.473B & 21.0\% \\
& Hybrid & 33 & 3840 & 10496 & 6.455B & 21.2\% \\
\bottomrule
\end{tabular}
\end{table}

\paragraph{Main on-policy recovery runs.}
Training uses the mixed math/code/open-ended research corpus of
Section~\ref{sec:setup} with rollout group $8$, train batch $64$, maximum
prompt and response lengths of $4096$ tokens each, rollout temperature
$0.8$, and learning rate $2\times10^{-6}$. The objective is the
top-$100$+tail generalized JSD with $\alpha=0.5$ and importance weights
clipped at $2.0$; no policy-gradient loss is used. The main schedule is
$710$ steps; the main-table checkpoint is step $709$.

\paragraph{Baselines and export.}
The offline baselines (SFT w/o KD, SeqKD, KD) train on the same prompt
distribution with the method-specific fixed responses of
Section~\ref{sec:setup}, using sequence length $8192$, batch $64$,
learning rate $2\times10^{-6}$ for $710$ steps; KD uses
forward KL at temperature $1.0$ with no hard-label cross-entropy term.
The matched Math+Code signal-control comparison trains PPO, GRPO, and
\shortopd\ on one deterministic shuffle of $29{,}585$ prompts:
$7{,}473$ GSM8K, $7{,}500$ MATH, $14{,}139$ OpenCodeInstruct, and $473$
MBPP train-side examples. APPS is not used. Each code row retains exactly
one executable test hidden from the policy prompt; its reference solution
must pass that test in two independent runs of the training sandbox before
the row is admitted. PPO and GRPO receive this binary verifier reward,
whereas \shortopd\ consumes only the shared prompt and the frozen teacher's
token distributions. Every method uses no-thinking chat templates, batch
$64$, $8$ rollouts per prompt, temperature $0.7$, and $2048$-token prompt
and response ceilings. The nominal ceiling count is $463$ batches; the
full-batch dataloader exhausts after optimizer step $462$, whose HF actor
export is evaluated for every method and scale. PPO uses GAE with a critic
initialized from the compressed student (learning rate $10^{-5}$), while
GRPO uses group-relative advantages. Full actor/critic state is retained
separately from the HF actor export.

\paragraph{Efficiency runs and controller.}
The \shortopd\ and fixed-horizon efficiency runs reuse the corpus,
objective, batch size, rollout count, and learning rate above; they
differ only in rollout temperature ($0.7$) and response ceiling ($2048$
for \shortopd\ and fixed $H{=}2048$; $8192$ for fixed $H{=}8192$). These
two $H{\leq}2048$ runs also provide the training trajectories shown in
Figures~\ref{fig:intro-prune-rep}(b), \ref{fig:shortopd-traj},
and~\ref{fig:shortopd-efficiency}(a). The controller
checkpoints its current budget and all EMA state across restarts.
Table~\ref{tab:controller-config} lists the detector and controller
hyperparameters; symbols follow Algorithm~\ref{alg:terminal-loop}.

\begin{table}[h!]
\centering
\small
\renewcommand{\arraystretch}{1.12}
\setlength{\tabcolsep}{6pt}
\caption{Detector and \shortopd\ controller hyperparameters.}
\label{tab:controller-config}
\begin{tabular}{@{}lc@{}}
\toprule
Parameter & Value \\
\midrule
Suffix window $W$ & $512$ \\
Candidate periods $p$ & $1$--$10$ \\
Terminal anchor / agreement $(A,\eta)$ & $(32, 0.90)$ \\
Minimum cycles / tail $(C,L_{\min})$ & $(3, 64)$ \\
Severe-loop tail & $\geq 128$ tokens or $30\%$ of suffix \\
Confirmation / baseline window & $32$ / $64$ tokens \\
OPD-loss threshold & $\max(0.05,\,0.25\,\bar d_{\mathrm{base}})$ \\
Teacher-NLL threshold & $\max(0.5,\,0.25\,\bar u_{\mathrm{base}})$ \\
Rate thresholds $(\rho_{\mathrm{low}},\rho_{\mathrm{high}},\tau)$ & $(0.20, 0.45, 0.10)$ \\
Statistic / budget EMA decays & $0.7$ / $0.7$ \\
Margin $\lambda$ / growth $\gamma_{\uparrow}$ & $1.15$ / $1.25$ \\
Budget rounding & multiples of $16$ \\
Initial budget / bounds & $2048$ / $[1024, 2048]$ \\
\bottomrule
\end{tabular}
\end{table}

\FloatBarrier

Because low loss participates in defining effective length, it never opens
the shrink gate: only the independent repetition-rate threshold can do so.
The probe does not mask the loss; training always applies dense OPD to the
tokens actually generated.

\section{Terminal periodic-loop detector}
\label{app:loop-detector}
Algorithm~\ref{alg:terminal-loop} gives the token-only detector used by
the controller. It searches only the last $W{=}512$ valid response tokens
and tests periods $p\in\{1,\ldots,10\}$. For a given period, each token is
compared with the token $p$ positions earlier. This shifted comparison does
not require the response to end on a cycle boundary, so a rollout truncated
part-way through its final cycle is still detected. Requiring high agreement
at the raw response end prevents repeated material followed by a clean ending
from firing the controller.

\begin{algorithm}[h!]
\caption{Severe terminal periodic-loop detection}
\label{alg:terminal-loop}
\small
\begin{algorithmic}[1]
\Require valid response tokens $y$; suffix window $W{=}512$; maximum period $P{=}10$;
terminal anchor $A{=}32$; agreement $\eta{=}0.9$; minimum cycles $C{=}3$;
minimum tail $L_{\min}{=}64$; severe length $L_{\mathrm{sev}}{=}128$;
severe fraction $\phi{=}0.30$
\State $z\gets\mathrm{Last}_{W}(y)$; $n\gets|z|$; $\mathcal C\gets\varnothing$
\For{$p=1,\ldots,\min(P,n-1)$}
  \State $c^{(p)}_k\gets\mathbb{1}[z_{p+k}=z_k]$ for $k=1,\ldots,n-p$
  \If{$\operatorname{Mean}(\mathrm{Last}_{A}(c^{(p)}))<\eta$}
    \State \textbf{continue} \Comment{the periodic pattern does not reach the response end}
  \EndIf
  \State $k_p\gets\max\{k:\operatorname{Mean}(\mathrm{Last}_{k}(c^{(p)}))\geq\eta\}$
  \State $L_p\gets k_p+p$; $u_p\gets\operatorname{Mean}(\mathrm{Last}_{k_p}(c^{(p)}))$
  \If{$L_p\geq L_{\min}$ \textbf{and} $L_p/p\geq C$
      \textbf{and} $(L_p\geq L_{\mathrm{sev}}$ \textbf{or} $L_p/n\geq\phi)$}
    \State add $(L_p,u_p,p)$ to $\mathcal C$
  \EndIf
\EndFor
\If{$\mathcal C=\varnothing$}
  \State \Return \textsc{not-severe}, $|y|$
\EndIf
\State $(L^*,u^*,p^*)\gets\arg\max_{(L,u,p)\in\mathcal C}(L,u,-p)$
\State \Return \textsc{severe}, $|y|-L^*$, $p^*$
\end{algorithmic}
\end{algorithm}

The lexicographic choice prefers the longest explained terminal tail, then
higher periodic agreement, then the shorter period. The detector alone
decides whether a rollout is severely repetitive. Its returned onset is the
structural effective-length candidate. We then search locally from that onset
for the first $32$-token window whose OPD loss and teacher NLL fall below their
absolute/relative thresholds; this refines the boundary but does not decide
whether repetition exists. If no loss-confirmed boundary is found, the
structural onset is retained. A rollout contributes to the \emph{clean
truncation rate} only if it fills $H_s$ and this detector does not mark it
severely repetitive.

\FloatBarrier

\section{Conditional OPD gradient at teacher--student agreement}
\label{app:gradient-derivation}
That a smooth divergence has zero gradient at its minimum is immediate;
the content of this appendix is the \emph{rate} at which the implemented
per-token gradient vanishes near teacher--student agreement, and that the
two implementation details -- top-$K$+tail aggregation and clipped
importance weighting -- preserve this stationary point.
Condition on one sampled prefix $(x,y_{<t})$ and treat that discrete state as
fixed during back-propagation, as in the implemented on-policy distillation
update. Let $p$ be the frozen teacher distribution and
$q=\operatorname{softmax}(z)$ the student distribution. For the canonical
forward-KL token loss $\ell_{\mathrm{KL}}=\mathrm{KL}(p\|q)$, the standard
softmax-cross-entropy derivative gives
\begin{equation}
\frac{\partial \ell_{\mathrm{KL}}}{\partial z_j}=q_j-p_j,
\qquad
\nabla_\theta\ell_{\mathrm{KL}}
=J_{z,\theta}^{\top}(q-p).
\label{eq:app-forward-kl-grad}
\end{equation}
The conditional gradient therefore vanishes as the student distribution
approaches the teacher distribution.

Our implementation uses generalized JSD rather than forward KL. For
$m=\alpha p+(1-\alpha)q$, the generalized Jensen--Shannon divergence is
\begin{equation}
D_\alpha(p,q)
=\alpha\sum_i p_i\log\frac{p_i}{m_i}
 +(1-\alpha)\sum_i q_i\log\frac{q_i}{m_i}.
\label{eq:app-gjsd}
\end{equation}
Because $\partial m_i/\partial q_i=1-\alpha$, differentiating both KL
terms makes their non-logarithmic terms cancel:
\begin{align}
\frac{\partial D_\alpha}{\partial q_i}
&=-\frac{\alpha(1-\alpha)p_i}{m_i}
 +(1-\alpha)\!\left(
 \log\frac{q_i}{m_i}+1-\frac{(1-\alpha)q_i}{m_i}
 \right) \notag\\
&=(1-\alpha)\log\frac{q_i}{m_i}.
\label{eq:app-gjsd-qgrad}
\end{align}
The softmax Jacobian is
$\partial q_i/\partial z_j=q_i(\mathbb{1}[i=j]-q_j)$. Applying the chain
rule therefore gives
\begin{align}
\frac{\partial D_\alpha}{\partial z_j}
&=(1-\alpha)\sum_i \log\frac{q_i}{m_i}
 q_i(\mathbb{1}[i=j]-q_j) \notag\\
&=(1-\alpha)q_j\left[
\log\frac{q_j}{m_j}-\sum_iq_i\log\frac{q_i}{m_i}
\right].
\label{eq:app-gjsd-zgrad}
\end{align}
At $q=p$, we also have $m=p$, so every logarithm in
Equation~\eqref{eq:app-gjsd-zgrad} is zero and the conditional logit
gradient is exactly zero. More locally, writing $q=p+\delta$ with
$\sum_i\delta_i=0$ and assuming $p_i>0$ on the retained support,
\begin{equation}
\log\frac{q_i}{m_i}
=\alpha\frac{\delta_i}{p_i}+O(\delta_i^2/p_i^2),
\qquad
\frac{\partial D_\alpha}{\partial z_j}
=\alpha(1-\alpha)\delta_j+O(\|\delta\|^2/\min_i p_i),
\label{eq:app-gjsd-local}
\end{equation}
so the logit gradient vanishes linearly as the two distributions agree
(at $\alpha{=}0.5$, with slope $1/4$ of the forward-KL gradient
$q_j-p_j=\delta_j$).
For model parameters $\theta$,
$\nabla_\theta D_\alpha=J_{z,\theta}^{\top}\nabla_zD_\alpha$; hence the
parameter gradient also tends to zero whenever the local logit Jacobian is
bounded.

Our implementation evaluates Equation~\eqref{eq:app-gjsd} after mapping the
vocabulary distribution into the teacher top-$100$ bins plus one aggregated
tail bin. This mapping is differentiable, and the analysis above applies
verbatim to the binned distributions, so the gradient vanishes already when
the \emph{binned} distributions agree -- a strictly weaker condition than
full-vocabulary agreement, since differences inside the tail bin are
invisible to the loss. This is the relevant condition for the loop states
of Section~\ref{sec:intro}: teacher and student need only agree at this
coarse granularity for the update to carry no signal. The clipped importance multiplier is bounded by
construction; at exact agreement, both the divergence and its gradient are
zero, so multiplying by this weight also preserves the stationary point.
This is a conditional, per-token statement rather than a claim that every
small scalar loss implies a small full-objective gradient. Empirically, the
low JSD measures teacher--student agreement on loop states, while the
separately reported low teacher NLL shows that the teacher itself assigns
high probability to continuing the observed loop.

\section{Evaluation details}
\label{app:eval}
GSM8K uses the full test split ($1{,}319$ examples) and exact/normalized answer
matching. MATH-500 uses boxed-answer extraction and normalization. The 4B
Instruct main results use no-thinking decoding with a $512$-token response
cap on GSM8K. MATH-500 uses a $16{,}384$-token response cap, temperature
$0.7$, top-$p=0.8$, top-$k=20$, and minimum-$p=0$; repetition stopping
only truncates detected degenerate loops. The 8B no-thinking MATH-500
column is re-evaluated with this same $16{,}384$-token protocol. Its
remaining cross-domain results use temperature $1$, top-$p=0.95$,
top-$k=64$, and response caps of $512$ for GSM8K, $2048$ for code, and
$768$ for open-ended tasks. Recovered Thinking checkpoints use greedy
decoding with an $8192$-token cap and remove the thinking trace before
scoring. Each row contains one response per test example. HumanEval
($164$ examples) and sanitized MBPP ($257$ examples) are evaluated with execution
\textsc{pass}@$1$ in a subprocess sandbox. Alpaca, QA, Summarization, and
MT-Bench~\citep{zheng2023mtbench} use fixed EAGLE-style prompt sets and are judged
on a $1$--$10$ scale by GPT-5.5, which is
neither the teacher nor a Qwen model.

The width/hybrid and matched Math+Code comparisons use a unified no-thinking
rerun. GSM8K and the seven non-MATH tasks use greedy decoding, with response
caps of $512$ for GSM8K and $768$ for code and open-ended generation.
MATH-500 is run separately with the $16{,}384$-token sampled protocol above.
For the width/hybrid comparison, multiple-choice evaluation applies each model's
chat template with thinking explicitly disabled, zero-shot MMLU, maximum
sequence length $4096$, and the mean log-probability of the candidate answer
letter after a fixed assistant \texttt{ANSWER:} prefix. The matched Math+Code
comparison retains the otherwise identical plain-prompt protocol.

\begin{table}[h!]
\centering
\scriptsize
\renewcommand{\arraystretch}{1.40}
\setlength{\tabcolsep}{2.7pt}
\caption{Remaining full-evaluation results for the matched Math+Code
comparison in Table~\ref{tab:rlvr-opd}. Open-ended columns are judge scores
on the $1$--$10$ scale; multiple-choice columns and MC Avg are accuracy
percentages.}
\label{tab:rlvr-full-eval}
\begin{tabular}{llccccccccc}
\toprule
Model & Method & Alpaca & QA & Sum & MT-B & ARC-C & HellaSwag & MMLU & WinoG. & MC Avg \\
\midrule
\multirow{3}{*}{\shortstack[l]{Qwen3-4B\\Instruct}}
& PPO & 1.03 & 1.00 & 1.01 & 1.04 & 54.01 & 44.39 & 42.99 & 51.30 & 48.17 \\
& GRPO & 1.05 & 1.01 & 1.06 & 1.01 & 44.11 & 33.72 & 38.40 & 48.86 & 41.27 \\
& \cellcolor{gray!25}\shortopd\ (ours)
& \cellcolor{gray!25}\best{2.81} & \cellcolor{gray!25}\best{1.41}
& \cellcolor{gray!25}\best{5.39} & \cellcolor{gray!25}\best{3.22}
& \cellcolor{gray!25}\best{70.14} & \cellcolor{gray!25}\best{63.39}
& \cellcolor{gray!25}\best{54.33} & \cellcolor{gray!25}\best{58.64}
& \cellcolor{gray!25}\best{61.63} \\
\midrule
\multirow{3}{*}{\shortstack[l]{Qwen3-8B\\No-thinking}}
& PPO & 1.14 & 1.03 & 1.15 & 1.11 & 77.65 & 55.85 & 54.46 & 61.25 & 62.30 \\
& GRPO & 1.04 & 1.04 & 1.01 & 1.02 & 74.57 & 50.90 & 51.60 & 59.35 & 59.11 \\
& \cellcolor{gray!25}\shortopd\ (ours)
& \cellcolor{gray!25}\best{2.86} & \cellcolor{gray!25}\best{1.38}
& \cellcolor{gray!25}\best{5.99} & \cellcolor{gray!25}\best{3.16}
& \cellcolor{gray!25}\best{88.65} & \cellcolor{gray!25}\best{80.29}
& \cellcolor{gray!25}\best{68.19} & \cellcolor{gray!25}\best{64.01}
& \cellcolor{gray!25}\best{75.29} \\
\bottomrule
\end{tabular}
\end{table}

\section{Repetition versus pruning depth: the full sweep}
\label{app:prunesweep}
\begin{figure}[h!]
\centering
\includegraphics[width=0.72\linewidth]{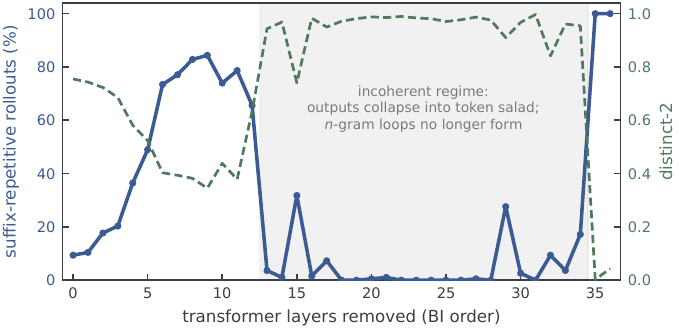}
\caption{Full pruning sweep on the $192$-prompt probe: suffix-repetition
rate (left axis) and distinct-$2$ diversity (right axis) versus the number
of BI-pruned layers. Repetition is the characteristic failure mode only in
the coherent regime ($k\!\leq\!12$); deeper pruning collapses generation
into incoherent text where loops no longer form.}
\label{fig:prune-sweep-full}
\end{figure}

Figure~\ref{fig:intro-prune-rep}(a) shows the suffix-repetition rate for
$0$--$12$ removed layers; Figure~\ref{fig:prune-sweep-full} extends the
sweep to all $36$ layers of Qwen3-4B-Instruct, with the distinct-$2$
diversity of the generations on the right axis. Three regimes emerge.
While the model remains a coherent generator ($k\!\leq\!12$), repetition
grows monotonically with removed depth and distinct-$2$ falls: the model
retains enough structure to sustain $n$-gram loops, and looping is its
dominant degeneration mode. Beyond $k\!\approx\!13$, outputs disintegrate
into incoherent token sequences: distinct-$2$ rebounds toward that of
random text and $n$-gram loops no longer form, so the measured repetition
rate collapses even though the model is far more damaged. At the extreme
($k\!\geq\!35$), the near-empty stack degenerates into trivial single-token
loops and the repetition rate saturates at $100\%$. The probe, generation
settings, and repetition detector are identical across all points
(Section~\ref{sec:shortopd-exp} and Appendix~\ref{app:train}).

\paragraph{Qualitative collapse regimes.}
Table~\ref{tab:collapse-cases} shows short excerpts from the same probe. The
$25\%$ compression point is still structured enough to produce recognizable
prefixes and repetitive suffixes; deeper compression often leaves the repetition
metric by entering an incoherent mixed-token regime instead.

\begin{table}[h!]
\centering
\scriptsize
\setlength{\tabcolsep}{4pt}
\renewcommand{\arraystretch}{1.2}
\caption{Probe excerpts illustrating the transition from normal generation to
repetition and then to incoherent collapse. Green indicates usable content;
red indicates repeated or incoherent content.}
\label{tab:collapse-cases}
\begin{tabular}{p{0.18\linewidth}p{0.76\linewidth}}
\toprule
Regime & Excerpt \\
\midrule
Uncompressed &
\textcolor{green!45!black}{\ttfamily Let's solve this step by step. We are told: Candice originally had 80 post-it notes... total post-its before placing on cups = post-its used + remaining.} \\
$25\%$ compressed, loop &
\textcolor{green!45!black}{\ttfamily Reason step step:}\ 
\textcolor{red!70!black}{\ttfamily step step step step solve step step step ... so.so.so. so.s.so.s ... ).").").").")} \\
Deeper compression, incoherent &
\textcolor{red!70!black}{\ttfamily Lonisolths\textasciitilde\ ``CCCCCCCC...'' Spain possessions ... JSON:!QQVV ... @@\#\# ... include\_app ...} \\
\bottomrule
\end{tabular}
\end{table}

\section{Multiple-choice recognition results}
\label{app:mc}
Although the main evaluation targets free-form generation, we also evaluate
held-out multiple-choice recovery on ARC-Challenge, HellaSwag, MMLU, and
WinoGrande. These benchmarks are not part of the recovery corpus. We collect
all multiple-choice tables in this appendix because recognition is a
complementary diagnostic rather than the paper's primary generation target.
The matched Math+Code comparison is reported in
Table~\ref{tab:rlvr-full-eval}; the two tables below cover width/hybrid and
depth-only pruning.

\subsection{Width and hybrid pruning}

We disable thinking and score candidate letters after a fixed assistant
\texttt{ANSWER:} prefix by mean token log-probability (zero-shot MMLU;
maximum length $4096$). Table~\ref{tab:structure-mc} shows that recognition
degrades less than generation: the pruned 4B-width and 8B-hybrid models retain
$51.14$ and $69.41$ MC Avg while reaching only $17.23$ and $16.17$ generation
Avg. \shortopd\ exceeds KD by $0.87$--$2.08$ MC points, smaller than its
$6.44$--$9.01$ generation gains in Table~\ref{tab:structure-extension}; thus
fixed-option recognition is not a sufficient proxy for generation.

\begin{table}[t!]
\centering
\scriptsize
\renewcommand{\arraystretch}{0.98}
\setlength{\tabcolsep}{4.2pt}
\caption{Held-out multiple-choice recognition after width-only and hybrid pruning.
Scores are accuracies in percent under chat-template candidate-letter scoring
with thinking disabled. MC Avg is the unweighted average of the four tasks.
Parenthetical percentages are the actual parameter reductions from
Table~\ref{tab:structure-config}.
Teacher and pruned rows use the same protocol as the recovered
checkpoints. Bold marks the best recovery value within each pruned
configuration, including ties.}
\label{tab:structure-mc}
\begin{tabular}{lllccccc}
\toprule
Model & Structure & Recovery & ARC-C & HellaSwag & MMLU & WinoG. & MC Avg \\
\midrule
\multirow{7}{*}{\shortstack[l]{Qwen3-4B\\Instruct}}
& Dense & Teacher & 90.53 & 80.29 & 70.87 & 68.11 & 77.45 \\
\cmidrule(lr){2-8}
& \multirow{3}{*}{Width (29.5\%)} & Pruned & 63.91 & 41.77 & 46.08 & 52.80 & 51.14 \\
& & KD & 73.81 & 58.39 & 54.76 & \best{58.88} & 61.46 \\
& & \shortopd\ (ours) & \best{76.62} & \best{63.25} & \best{55.42} & \best{58.88} & \best{63.54} \\
\cmidrule(lr){2-8}
& \multirow{3}{*}{Hybrid (29.4\%)} & Pruned & 74.40 & 58.45 & 53.13 & 54.85 & 60.21 \\
& & KD & 82.68 & 66.52 & 59.27 & \best{60.54} & 67.25 \\
& & \shortopd\ (ours) & \best{83.53} & \best{69.63} & \best{60.40} & 59.91 & \best{68.37} \\
\midrule
\multirow{7}{*}{\shortstack[l]{Qwen3-8B\\No-thinking}}
& Dense & Teacher & 91.72 & 83.35 & 72.07 & 68.67 & 78.95 \\
\cmidrule(lr){2-8}
& \multirow{3}{*}{Width (21.0\%)} & Pruned & 71.42 & 60.73 & 53.23 & 54.85 & 60.06 \\
& & KD & 79.86 & 68.01 & 59.98 & 60.46 & 67.08 \\
& & \shortopd\ (ours) & \best{80.46} & \best{70.58} & \best{60.90} & \best{60.85} & \best{68.20} \\
\cmidrule(lr){2-8}
& \multirow{3}{*}{Hybrid (21.2\%)} & Pruned & 82.42 & 72.62 & 61.26 & 61.33 & 69.41 \\
& & KD & 84.04 & 73.59 & 63.36 & \best{63.85} & 71.21 \\
& & \shortopd\ (ours) & \best{84.56} & \best{75.77} & \best{64.12} & \best{63.85} & \best{72.08} \\
\bottomrule
\end{tabular}
\end{table}

\subsection{Depth-pruning sanity check}

Table~\ref{tab:mc-sanity} uses candidate log-likelihood scoring on
Qwen3-4B-Instruct. SFT-init \shortopd\ starts from the one-epoch SFT w/o KD
checkpoint rather than the pruned student. This off-policy recognition task
favors SFT and KD ($74.9$/$74.4$ Avg), although they trail on generation in
Table~\ref{tab:main-results}. One-epoch \shortopd\ reaches $67.6$ MC Avg,
$22$ points above Pruned, and rises to $71.6$ with three epochs. Combining
the complementary signals, SFT-init \shortopd\ reaches $76.8$, matching the
Teacher at $76.7$.

\begin{table}[h!]
\centering
\small
\renewcommand{\arraystretch}{1.15}
\setlength{\tabcolsep}{4.2pt}
\caption{Held-out multiple-choice recovery for the $25\%$-pruned
Qwen3-4B-Instruct, scored by candidate log-likelihood over the answer
options. Scores are accuracies in percent.}
\label{tab:mc-sanity}
\begin{tabular}{lccccc}
\toprule
Method & ARC-C & HellaSwag & MMLU & WinoG. & Avg \\
\midrule
Teacher                   & 89.93 & 80.60 & 70.93 & 65.19 & 76.66 \\
Pruned ($-25\%$)          & 49.49 & 41.67 & 41.65 & 49.80 & 45.65 \\
SFT w/o KD                & 89.33 & 76.89 & 69.09 & 64.09 & 74.85 \\
SeqKD                     & 83.36 & 72.65 & 64.19 & 63.06 & 70.82 \\
KD             & 87.71 & 77.13 & 67.59 & 65.04 & 74.37 \\
\shortopd\ (1 epoch)      & 79.52 & 68.89 & 60.70 & 61.33 & 67.61 \\
\shortopd\ (3 epochs)     & 84.47 & 75.59 & 64.59 & 61.72 & 71.59 \\
SFT-init \shortopd\       & \best{89.93} & \best{79.67} & \best{69.99} & \best{67.64} & \best{76.81} \\
\bottomrule
\end{tabular}
\end{table}

\end{document}